\documentclass{article}

\usepackage{PRIMEarxiv}

\usepackage[utf8]{inputenc} 
\usepackage[T1]{fontenc}    
\usepackage{hyperref}       
\usepackage{url}            
\usepackage{booktabs}       
\usepackage{amsfonts}       
\usepackage{nicefrac}       
\usepackage{microtype}      
\usepackage{lipsum}
\usepackage{fancyhdr}       
\usepackage{graphicx}       
\graphicspath{{media/}}     
\usepackage{amsmath}
\usepackage{multirow}
\usepackage{caption}
\title{Ultra-High-Resolution Image Synthesis with Pyramid Diffusion Model
\thanks{
\textbf{Preprint Version. Finer version and more results are coming soon. Code available from: \href{https://github.com/RANDO11199/pyramid-diffusion}{here}  }} 
}

\author{
  Jiajie Yang\\
  Personal Researcher \\
  \texttt{jiajie.y@wustl.edu} \\
}

\begin{document}

\maketitle
\begin{center}
\centering
\captionsetup{type=figure}
\includegraphics[width=\textwidth,height=.5\textwidth]{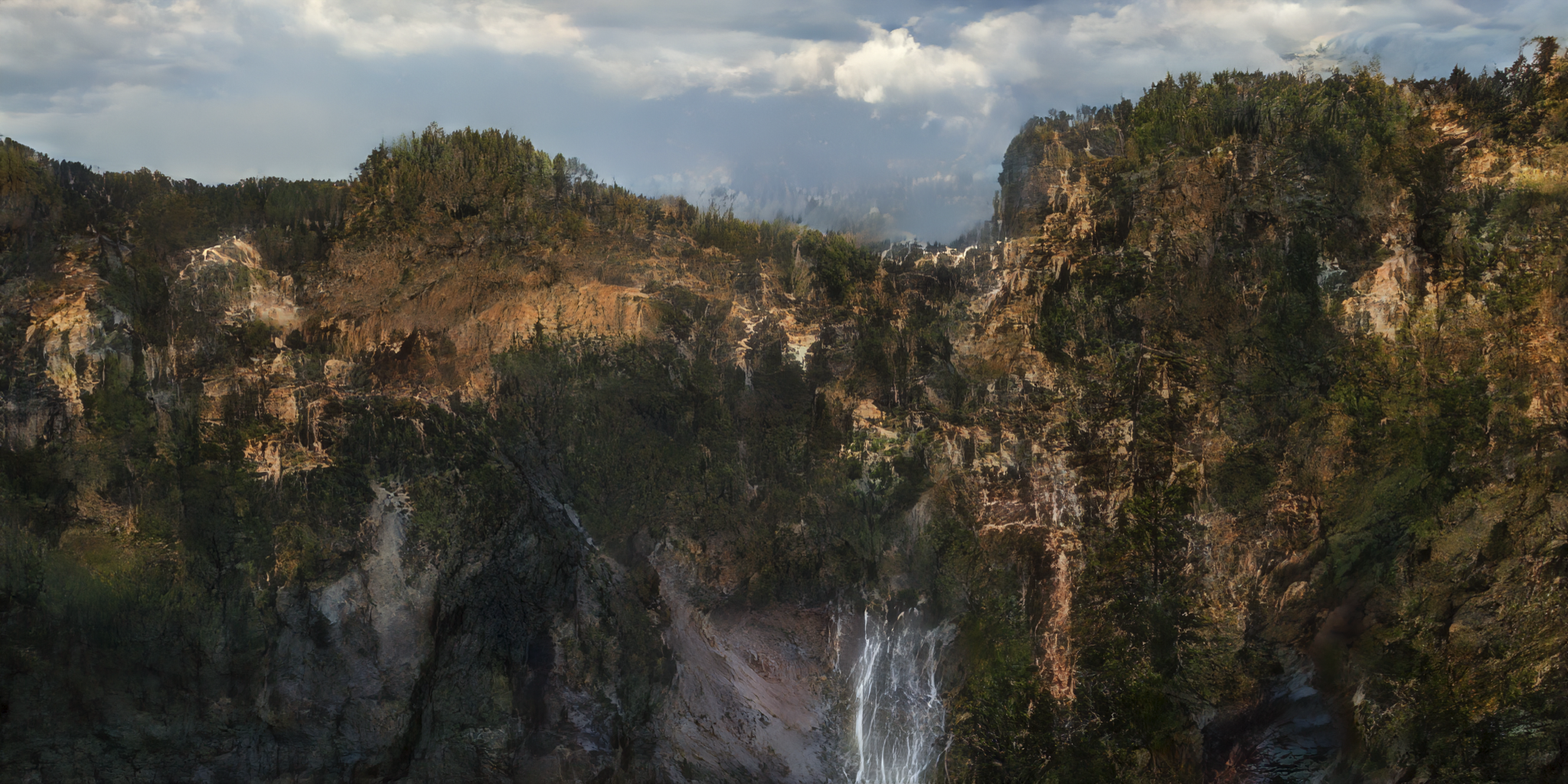}
\captionof{figure}{Pyramid Diffusion Model could generate really-high-resolution images unconditionally. The displayed image contains 2048*1024 pixels.}
\end{center}
\begin{abstract}
We introduce the Pyramid Diffusion Model (PDM), a novel architecture designed for ultra-high-resolution image synthesis. PDM utilizes a pyramid latent representation, providing a broader design space that enables more flexible, structured, and efficient perceptual compression which enable AutoEncoder and Network of Diffusion to equip branches and deeper layers. To enhance PDM's capabilities for generative tasks, we propose the integration of Spatial-Channel Attention and Res-Skip Connection, along with the utilization of Spectral Norm and Decreasing Dropout Strategy for the Diffusion Network and AutoEncoder. In summary, PDM achieves the synthesis of images with a 2K resolution for the first time, demonstrated on two new datasets comprising images of sizes 2048x2048 pixels and 2048x1024 pixels respectively. We believe that this work offers an alternative approach to designing scalable image generative models, while also providing incremental reinforcement for existing frameworks.
\end{abstract}


\section{Introduction}
\label{sec:intro}

Diffusion model has recently achieved significant success in generative tasks. Among the various diffusion models, Latent Diffusion Model is widely adopted in industry for its ability to strike a balance between perceptual quality and computational efficiency, as well as its flexible conditioning methods for various tasks. For densely conditioned tasks such as super-resolution, inpainting, and semantic synthesis, Latent Diffusion model \cite{ldm} has demonstrated the capability to generate large, consistent images of 1024x1024 pixels. Subsequently, several notable research efforts have focused on improving Latent Diffusion Models \cite{podell2023sdxl} \cite{pernias2023wuerstchen}, successfully generating images of up to 1024x1024 pixels conditioned on text prompts. Additionally, there have been intriguing attempts to combine latent diffusion with scalable network architectures for high-resolution image synthesis \cite{bao2023ViTLDM} \cite{peebles2023scalable}.

While many existing works focus on scaling up the diffusion model with more stages or larger networks, to the best of our knowledge, no research has focused on the latent representation. In general case, a single latent with a spatial compression rate of f-4 or f-8 is the default choice. However, such an approach could be limiting. Firstly, single latent restricts the design space of the AutoEncoder in the Latent Diffusion Model (LDM). To achieve acceptable reconstruction quality, the AutoEncoder needs to remain shallow in terms of resolution level and network depth. However, it's widely observed that deeper networks generally yield better performance \cite{2019propagation}\cite{eldan2016power}, as deeper networks can learn richer spatial structures with different receptive fields, enabling the network to learn features at various levels of abstraction. Additionally, it is recommended to have wider networks \cite{zagoruyko2017wide}. Secondly, a stronger AutoEncoder with more parameters also alleviates the pressure on the Diffusion Process to generate high-quality images, as the AutoEncoder is only evaluated once. Thirdly, features in deep layers with lower resolution are simpler to extend in a channel-wise manner, which is thought to be helpful and cost-effective for image synthesis \cite{hoogeboom2023simple}. In these aspects, our framework takes advantage of deep and wide neural networks as well as a larger number of channels. Additionally, we assume that a single latent might force the network to produce aliasing features because the network may be compelled to introduce new content as soon as possible and remember aliasing information in all layers, as only one single latent is being encoded containing most of the necessary information to reconstruct the image from the bottom of the network gradually, without proper separation of image's elements, such as layout, color and texture.

\begin{figure}[htb]
  \centering
  \includegraphics[height=6.5cm]{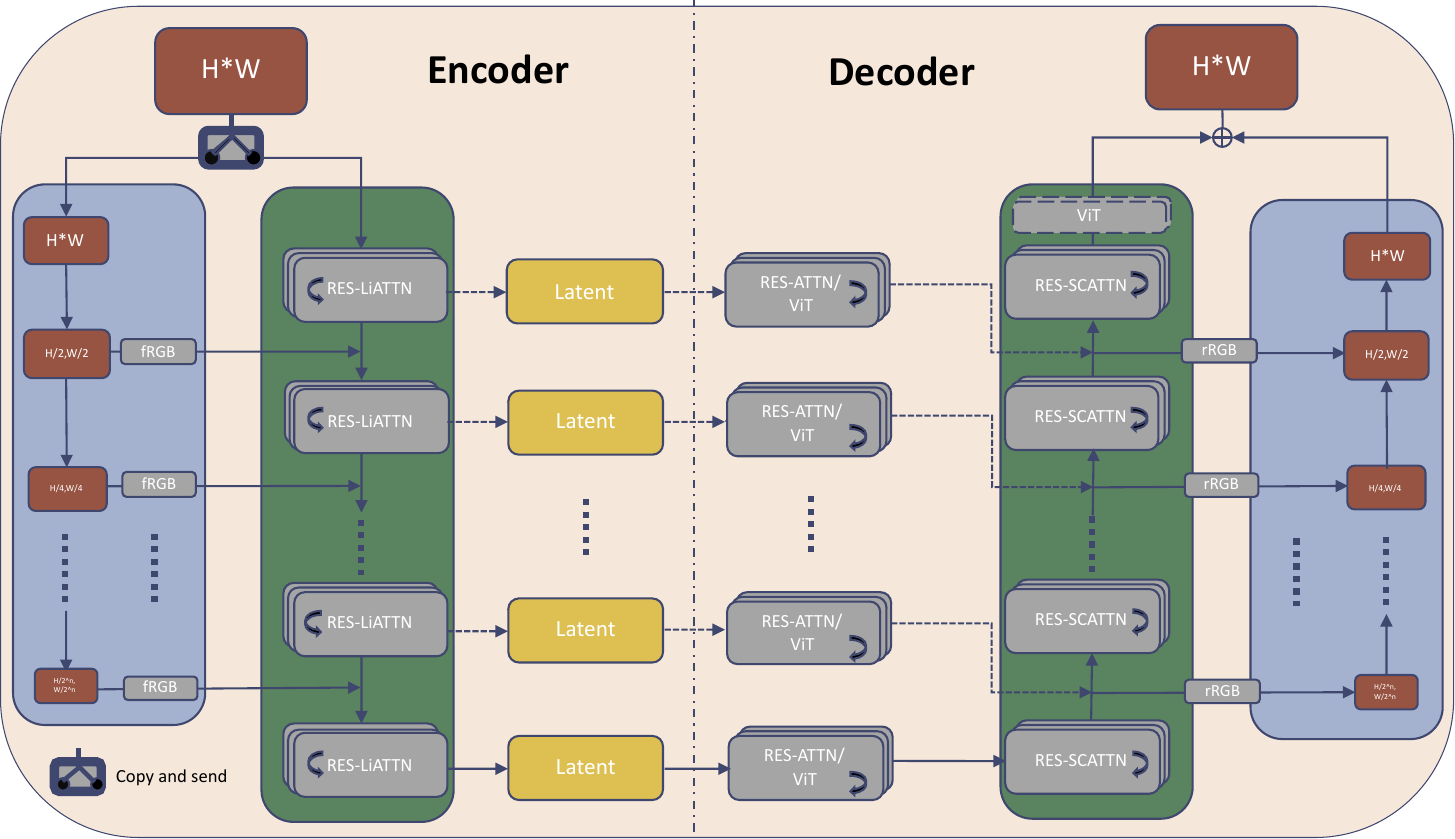}
  \caption{The main architecture of Pyramid AutoEncoder. The main architecture of the AutoEncoder consists of several components. The component enclosed in blue boxes represents the Input/Output Skip mechanism proposed in \cite{styleganplus}. The input image is sent to a stream of Input-Skip and down-sampling stream, where it is encoded into a latent representation. Subsequently, the latent representation is forwarded to the Decoder for image reconstruction. It is worth noting that our AutoEncoder design assigns different importance to the Encoder and Decoder. The Encoder adopts lightweight components, heavy regulation, and a non-branch design, while the Decoder utilizes heavyweight components, light regulation, and a branch design.}
  \label{fig:autoencoder}
\end{figure}

To overcome the limitations on network design in the latent diffusion model, we propose extending the single latent introduced in \cite{ldm} to a pyramid latent structure with varied resolutions. Specifically, as depicted in Fig. \ref{fig:autoencoder}, we now pop out latents from the Encoder at different resolutions alongside the original single latent and extend the Decoder with branches to process the corresponding latents. For instance, instead of the usual resolutions like 32x32 or 64x64 pixels, we down-sample an image with 256x256 pixels to features with resolutions as low as 1x1 pixels. Subsequently, we extract latents from the bottom with resolutions of 1x1, 4x4, 8x8, and 16x16. Pyramid latent structures provide multiple sources of information with varying levels of abstraction, allowing the AutoEncoder to progressively reconstruct the image from semantic concepts to details such as texture and edges explicitly. This is in contrast to an AutoEncoder with a single latent, where the layers in the network focus on mixed elements of the image. The decomposition of the latent representation also enables flexible and independent refinement of the image with branches in the Decoder. With these improvements, the total compression rate of our AutoEncoder can be as high as 1:256. For instance, we reconstruct images from CelebA-1024 with dimensions of 1024x1024x3 using pyramid latents with dimensions of 16x16x32 + 8x8x64 (see Table \ref{table:autoencoder} and reconstruction samples in Fig. \ref{sampleceleba}). Lastly, only slight modifications are applied to the UNet, where branches are added to the original UNet to accommodate pyramid latents.

Additionally, we introduce several new components into the neural network. First, we adopt skip connections \cite{styleganplus} and propose novel residual-skip connections for both the AutoEncoder and the Diffusion UNet. Second, we introduce Spatial-Channel Attention to replace the spatial self-attention conventionally used in the Diffusion UNet. We will discuss these components in more detail in Section \ref{Subsection:PDM}. Thirdly, we suggest phenomenon which we named as \textbf{Concept Aliasing} based on experiment observation and adopt Decreasing Dropout Strategy to relieve this problem, we will give a detailed discussion later at section.\ref{Subsection:PDM}. Lastly, we choose Rectified Flow \cite{liu2022rectflow} as our diffusion framework.

Bringing together these advancements, we introduce the Pyramid Diffusion Model (PyraDiffusion). In this paper, we succinctly achieve the following:

1. We replace the single latent in the Latent Diffusion Model with pyramid latent structures, enabling a more flexible design choice and facilitating the use of larger AutoEncoders for high-quality reconstructions with high compression rates.

2. We propose the Pyramid UNet, equipped with branches to model pyramid latent structures.

3. We introduce spatial-channel attention to accommodate rapid changes in pixels and channels within the UNet and AutoEncoder.

4. We suggest a res-skip architecture, which combines residual connections with input/output-skip connections from \cite{styleganplus}.

5. We curate two datasets, each with dimensions of 2048x2048 pixels and 2048x1024 pixels respectively.

6. We propose a decreasing dropout strategy for training the Diffusion UNet and AutoEncoder.

With these enhancements, we achieve to unconditionally produce images with a resolution of 2K for the first time, as well as  unconditionally generate images with dimensions of 1024x1024 pixels using the latent diffusion model for the first time

\section{Related Work}
\textbf{High Resolution Image Synthesis} High resolution image synthesis has been researching for long. The essentially high-dimensional nature incurs huge difficulty for generative model. Situation become worse and resolution of image increase. Generative Adversarial Networks[\cite{gan2014}] is one of the most successful generative frameworks in the era of deep learning. GAN allows one-step image sampling with high quality\cite{biggan}\cite{styleganplus}\ and achieve outstanding performance in many image datasets\cite{stylenat}\cite{styleganXL}\cite{projectedgan} in the sense of metrics such as FID\cite{fid}, IS Score\cite{ISScore} and precision\cite{precision}. Nevertheless, GANs suffer from stability problem during training\cite{unstableGan} and many works has been done to relieve such difficulties\cite{WGAN}\cite{WGAN-GP}. Many recent researches also suggested a relative lack of diversity compared to likelihood-based methods \cite{vqvae2}\cite{IDDPM}\cite{sparse-representation-gen}. Nonetheless, fast inference speed, latent space with good properties help GAN achieve good performance in many tasks, such as image editing \cite{stylegan}\cite{pan2023drag}\cite{ling2021editgan}. Comparing to GAN, likelihood-based models share the well-behaved optimization dynamics but find difficult to either comparable image quality\cite{VAE}\cite{realNVP}\cite{fvae} or inference speed\cite{dinh2014nice}\cite{pixelcnn}\cite{pixelrnn}. 

\textbf{ODE-and-SDE-based Model}. Recently, Diffusion/SDE-based model have obtaining great success\cite{ho2020ddpm}\cite{song2020score} and outperforming GAN in image synthesis in both variety and quality\cite{dhariwal2021diffusionbeatgan}, although many recent works prove Gan a useful tools in specific applications \cite{pan2023drag}\cite{vecGanpp}\cite{li2022supervisedGan}. Also, Diffusion framework succeeded to build large text-to-image models with excellent results \cite{nichol2022glide}\cite{ramesh2022hierarchicalt2i}. Diffusion also achieve great success in many domain, such as image inpainting \cite{ldm}, video \cite{xing2023diffusionvideosurvey} and image editing\cite{zhao2022egsde}. However, generally Diffusion/SDE model needs hundreds of expensive evaluations step on Neural Network and thus slow reference speeds. Many works aimed at improving the inference speed with best image quality, such as high order solver for PF-ODEs\cite{lu2022dpmsolver},systematic design space examination\cite{karras2022elucidating} and DDIM \cite{ddim}. Furthermore, by leveraging ODE-based model, one-step generation had been suggested\cite{song2023consistencymodel}\cite{liu2022rectflow}. While \cite{song2023consistencymodel} tried to match any points on the ODE trajectory to its origin and \cite{liu2022rectflow} chose to design a straight ode trajectory to match origins and noise, both methods achieve good results theoretically and empirically. In this paper, we choose \cite{liu2022rectflow} to fit the distribution of encoded latent representations as this method offer a intuitive yet simple implementation of Diffusion ODE and natural way to decrease reference steps. 

\textbf{Two-Stage Model}. Two-Stage model firstly encode the data into latent representation and afterward fit the distribution of encoded features with probabilistic framework combining with Deep Neural Network.  VAE \cite{VAE} showed the advantages of first learning a latent representations and then learning its distribution with VAE again. VQ-VAEs\cite{vqvae} and VQ-VAE2\cite{vqvae2} learn first a quantized latent representations and then fit the quantized latent with autoregressive models. \cite{ramesh2021zeroshot} improve this approach to achieve text-to-image generation. \cite{ldm} presents Latent Diffusion Model (LDM), and approach first learn discrete or continuous latent representations and model their distribution in Diffusion framework with Unet and convolution architecture and thus more gently to higher dimensional latent spaces compare to \cite{esser2020taming} \cite{ramesh2021zeroshot}. Also, \cite{podell2023sdxl} \cite{pernias2023wuerstchen} adopt multiple stage to refine latent produced by diffusion process. Our work further prevents the trade-offs between compression rate of encoded latent and number of trainable parameters. Our model also allow more granular design on level of compression rate and more efficient use of computation power. Also, our model are highly scalable due to the layered design and therefore a good choice for large text-to-image model\cite{podell2023sdxl}.We choose to separately learn our Two-Stage model \cite{sinha2021d2c} due to the limitation of video memory and avoid the difficut weighting between reconstruction and generative capabilities.
\section{Method}
\label{sec:method}
\subsection{Rectified Flow}
Given $X_{1}$ and $X_{0}$, Rectified Flow tries to modeling the evolutionary process $X_{1} \to X_{0}$ and follow the equation:
\begin{align}
    \frac{dX_{t}}{dt} = v_{t}(X_{t}),
\end{align}
where $t\in[0,1]$ and $v_{t}(X_{t})$ is instantaneous speed of continuous dynamical system.  Given $X_0\sim\pi_{0} $ and $X_{1} \sim \pi_{1}$, \cite{liu2022rectflow} assume the trajectory $\phi_{t}(X_{0},X_{1})$ to be straight line:
\begin{align}
    X_{t} = \phi_{t}(X_{0},X_{1}) = tX_{1} + (1-t)X_{0},
\end{align}
where $X_{t}$ is the linear interpolation of $X_{0}$  and  $X_{1}$. Therefore, the speed of the dynamical system at any time $T$:
\begin{align}
    \frac{dX_{t}}{dt} = v_{t}(X_{t}) = X_{1} - X_{0}.
\end{align}

However, the target distribution $X_{1}$ is not known, \cite{liu2022rectflow} leverage neural network to simulate this process who choose to minimize the Mean Square Error (MSE) between the approximated $\Tilde{v_{t}}$ and $(X_{1}- X_{0})$
\begin{align}
\label{Objective Function}
L_{RF} & = E_{t\sim U[0,1]}\left[ E_{X_{0}\sim noise,t} \left[\left(|| (X_{1} - X_{0}) - v(X_{t},t) ||^2 \right)\right]\right],
\end{align}
given $\phi_{t}(X_{0},X_{1})$ is reversible w.r.t. $X_{1}$ and using test function, \cite{liu2022rectflow} show equation (6) achieves the transportation between $\pi_{0}$ and $\pi_{1}$ with consistent dynamics.

\subsection{Latent Rectified Flow Models}

\textbf{Latent Diffusion Model} With trained Encoder $\mathcal{E}$ and Decoder $\mathcal{D}$, Latent Diffusion Model (LDM) obtain an efficient, low-dimensional latent representation $z_{t}=$ $\mathcal{E}$(X) where high-frequency and imperceptible details are abstracted away. \cite{ldm} then optimize the reweighted bound\cite{ho2020ddpm}:
\begin{align}
  L_{LDM} & = E_{\mathcal{E}(x),\epsilon\sim N(0,1),t} \left[||\epsilon - \epsilon_{\theta}(z_{t},t)||^{2}_{2}  \right]. 
\end{align}

\textbf{Latent Rectified Flow} Instead of fitting the trajectory between distribution $\pi_{0}(X_{0})$ and $\pi_{1}(X_{1})$ in high dimensional image space, we first encode image distribution into latent distribution $\mathcal{E}(X_{1})$. In latent space, we model the straight trajectory between $Z_{0}\sim N(0,1)$ and $Z_{1}\sim \pi_{1}(\mathcal{E}(X_{1}))$ :

\begin{align}
L_{LRF} & = E_{t\sim U[0,1]}\left[ E_{Z_{0}\sim noise,t} \left[\left(|| (Z_{1} - Z_{0}) - v(Z_{t},t) ||^2 \right)\right]\right],
\end{align}
where $Z_{t} = tZ_{1} + (1-t)Z_{0}$.

\subsection{Pyramid Diffusion Models}
Pyramid Latent Diffusion Models combines pyramid latent, several new architecture and regulation strategies together. For architecture, we propose Spatial-Channel Attention, Res-Skip Connection, pyramid AutoEncoder, pyramid UNet. For regulation strategies, we propose Spectral Norm and Decreasing Dropout Strategy.

\textbf{Pyramid Latent Diffusion} Pyramid Latent Diffusion Models adopt pyramid latent architecture and Rectified Flow as diffusion model. Correspondingly, the loss function become:
\begin{align}
L_{PDM} & = \sum_{i} E_{t\sim U[0,1]}\left[ E_{Z_{0}\sim noise,t} \left[\left(|| (Z_{1,i} - Z_{0,i}) - v(Z_{t,i},t) ||^2 \right)\right]\right],
\end{align}
where i refer to i-th latent in the pyramid and  $Z_{t,i} = tZ_{1,i} + (1-t)Z_{0,i}$. 
\label{Subsection:PDM}

\textbf{Spatial-Channel Attention} Different from both spatial attention and channel-wise attention, spatial-channel attention calculates attention across both pixels and channels, weighting the attention based on their respective importance in terms of scale.

\begin{figure}[htb]
  \centering
  \includegraphics[height=4.0cm]{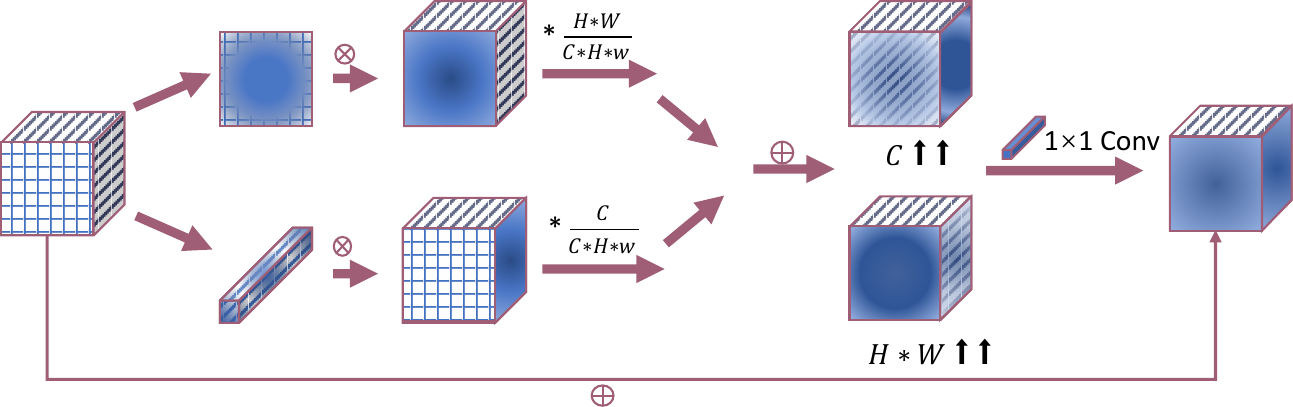}
  \caption{ The Spatial-Channel attention calculates spatial attention and channel attention concurrently. After obtaining the two weighted features, we re-weight these features with the ratio of importance measured by the scale of pixels or channels. Our Spatial-Channel attention offers the ability for self-adaptation to intense changes in resolution and channels as it takes care of both channel and image features simultaneously.
  }
  \label{fig:scattn}
\end{figure}

To be more specific, spatial-channel attention computes spatial and channel attention in parallel. Subsequently, to ensure a smooth transition between spatial attention and channel attention, we reweight features:
\begin{align}
\label{reweighted Attn}
Feature = \frac{H*W}{H*W+C} * Feature_{spatial} +  \frac{C}{H*W+C} * Feature_{channel},
\end{align}
where H*W is the number of pixels and C is number of channels (see Fig.\ref{fig:scattn}).

\begin{figure}[h]

\begin{minipage}[p]{0.5\textwidth}
  \centering
  \resizebox{\linewidth}{!}{
  \includegraphics[width=60mm]{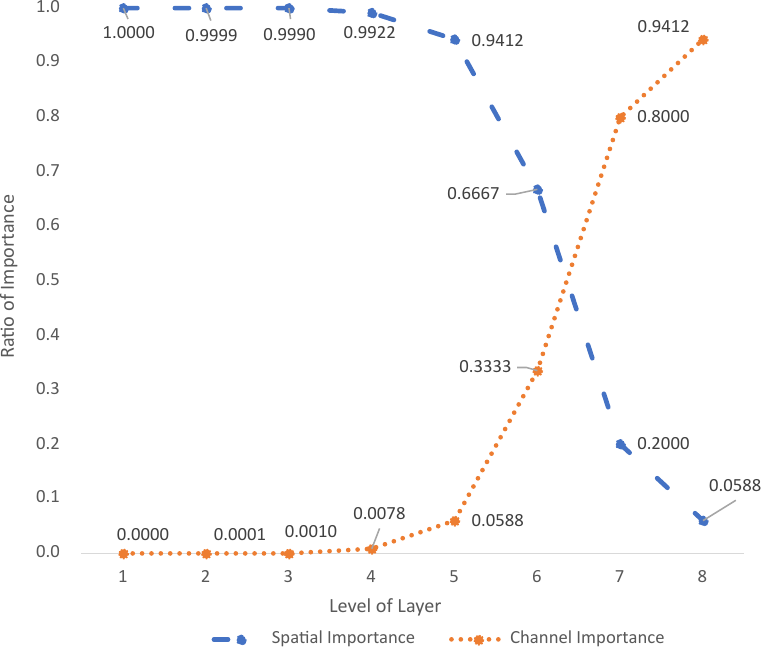}
  }
\end{minipage}%
\begin{minipage}[p]{0.5\textwidth}
  \centering
  \resizebox{\linewidth}{!}{
  \begin{tabular}{@{}l|l|l|l|l@{}}
    \toprule
    Level & Height & Channel & Spatial Weight  &  Channel Weight\\
    \midrule
    1 & 1024& 16& 1.0000&	     0.0000\\
    2 & 512 &	32 &	0.9999 &	0.0001\\
    3 & 256 &	64 &	0.9990 &	0.0010\\
    4 & 128 &	128 &	0.9922 &	0.0078\\
    5 & 64 &	256 &	0.9412 &	0.0588\\
    6 & 32 &	512 &	0.6667 &	0.3333\\
    7 & 16 &	1024 &	0.2000 &	0.8000\\
    8 & 8 &	1024&	0.0588 &	0.9412\\
  \bottomrule
  \end{tabular}
  }
\end{minipage}
\caption{\textbf{(Left)} The figure illustrates the ratio of importance of spatial attention and channel attention across different layers in the neural network. Following the conventional CNN design, down-sampling is accompanied by an increase in channel dimensions. A noticeable fluctuation in importance between channel and spatial attention is observed in deeper layers of the network compared to the initial layers. \textbf{(Right)} The table presents the data corresponding to the left figure.} 
\label{fig:ratirochannelpixel}
\end{figure}

In contrast to the utilization of either spatial attention or channel attention in isolation, spatial-channel attention adeptly manages both pixel and channel features. Consider a neural network with an initial channel count of 16 and an input image measuring 1024x1024 pixels. Also, as per convention, channel count is doubled while width and height are halved. Initially, the ratio of pixels to the sum of pixels and channels is approximately 1, justifying a focus on spatial relationships, which initially predominate. However, with network forwarding deeper, the significance of spatial and channel dimensions undergoes a gradual inversion (See Fig.\ref{fig:ratirochannelpixel}). For instance, when examining features with dimensions of 16x16 pixels and 1024 channels, the ratio of pixels to pixels plus channels diminishes to 0.2, highlighting the advantages of channel over pixel. In cases where this ratio approximates 0.5, such as when dimensions are set at height=32, width=32, and channel=1024, equitable differentiation between channels and pixels becomes more rational. By integrating both channel and spatial attention mechanisms, spatial-channel attention obviates the challenge associated with selecting between spatial and channel-wise attention mechanisms with different ratio of pixels to pixels plus channels and both pixels and channels is taking into consideration.

\begin{figure}[h]
  \centering
  \includegraphics[height=6.5cm]{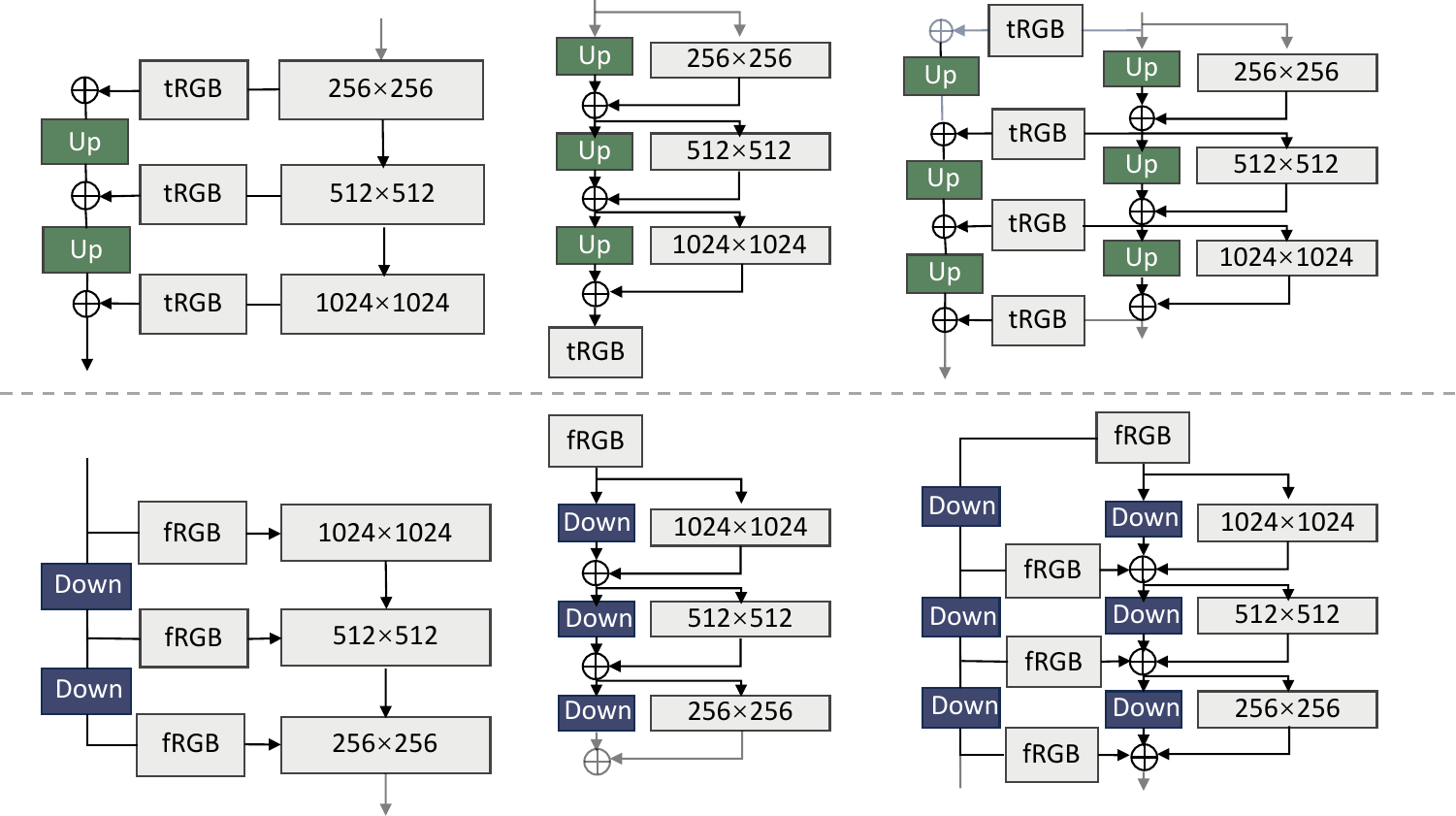}
  \caption{ The Residual and Skip Connection design is illustrated in the up-sampling stream (above the gray dotted line) and the down-sampling stream (below the gray dotted line). The images in the left column depict the structure of Input/Output skips, the middle column illustrates the structure of Residual nets, and the right column describes the structure of Res-Skip connections. We implement the \framebox{Up}-sampling and \framebox{Down}-sampling techniques as outlined in \cite{song2020score} in the Pyramid NCSNpp. Meanwhile, in the Decoder, we employ nearest neighbor interpolation for up-sampling, and in the Encoder, we use pooling (max-pooling or average pooling) for down-sampling. Additionally, \framebox{tRGB} and \framebox{fRGB} modules are utilized to convert between RGB and high-dimensional per-pixel data.
  }
  \label{fig:resskip}
\end{figure}

\textbf{Res-Skip Connection} Many previous studies have proven skip connections \cite{ronneberger2015unet, karnewar2020msggan} and residual networks \cite{gulrajani2017improvedWGANResNet} to be successful methods in the context of generative methods. Combining skip connection and residual networks, \cite{styleganplus} propose input/output skips structure (Figure \ref{fig:resskip}, column 1) and residual structure (Figure \ref{fig:resskip}, column 2). While both structures are proven to be effective in improving the quality of generated images separately in \cite{styleganplus}, the combinations of input/output skip and residual nets affect the final results differently across different datasets.

To be more elaborate, in the FFHQ dataset, the combination of output skips in the Generator and residual net in the Discriminator (instead of input skips) obtains a significant improvement compared to other combinations. Conversely, in the LSUN Car dataset, the residual structure excels over other combinations dramatically (FID 3.25, 3.19, 2.66 for Generator without any structure, Generator with output skips, and Generator with residual). 

Although \cite{styleganplus} indicates results on LSUN Car as a lone exception, it might also suggest a potential benefit of residual structure in some kinds of data given such a significant difference in FID score. Observing the natural cohesion of these two structures and experimental results in \cite{styleganplus}, we suggest a new structure named Res-Skip Connection (Figure \ref{fig:resskip}), which grasps both the key aspects of progressive growing \cite{karras2018progressive}—being able to let the network focus on low-resolution features and then slowly shift its attention to finer details—and the benefits of residual architecture with gradient and feature preserving .

\textbf{Spectral Norm}  We choose to add Spectral Norm in UNet architecture for three reasons:

    Stable Sampling. For a well-trained network, achieving local consistency allows larger error tolerance for velocity estimator, which implies that the velocity estimator, $v(\Tilde{x}_{t},t)$ should approximate $v(x_{t},t)$ in neighborhood, where $\Tilde{x}_{t}$ represents the estimated position at time $t$ obtained by solving the initial value problem of the ODE equation, and $x_{t}$ denotes the exact position on the trajectory between $X_{0}$ and $X_{1}$.
    
    Stable Encoder. To enhance the stability of the encoder to small perturbation, we incorporate spectral norm into both encoders. For an encoder $\mathcal{E}(x)$ equipped with spectral norm, we ensure that $||\mathcal{E}(X_{j}) - \mathcal{E}(X_{i})|| \leq ||X_{i}-X_{j}||$, signifying that given two similar images, the encoder would encode similar latent code with differences bounded by $||X_{i} - X_{j}||$ and thus produce more compact latent code as well as be stable to perturbation.
    
    Balance between regulaton and model capability. Comparing this approach to weight normalization (e.g., L2 norm) aimed at enhancing the generalizability \cite{yoshida2017spectral} of the model, the Lipschitz constraint of a linear operator is solely determined by the maximum singular value, implying that weight matrix only shrinks in one direction. Thus, it allows the parameter matrix to utilize as many features as possible with some degree of regulation while enabling desensitization of network by adhering to the local 1-Lipschitz constraint \cite{miyato2018spectral}.
    
\textbf{On AutoEncoder} Latent Diffusion offers two options for AutoEncoders: Variational AutoEncoder \cite{VAE} and Vector-Quantized Variational AutoEncoder \cite{vqvae}. Building upon these designs, we have made three main modifications to the AutoEncoder.

Firstly, we have adopted a lightweight encoder and a heavyweight decoder. The basic module of the encoder comprises a ResBlock and Linear Attention, while the basic module of the decoder comprises a ResBlock and Spatial-Channel Attention. Additionally, each module in the decoder has one more block than in the encoder. Furthermore, we have extended the decoder with branches consisting of ResBlock + Spatial-Channel Attention or Vision Transformer (ViT) to process the latent representation before the backbone.

Secondly, we have added an input-skip module in the encoder and an output-skip module in the decoder \cite{karras2020analyzing}.

Thirdly, we have introduced Spectral Norm in the encoder, while the decoder remains regulation-free.

\begin{figure}[htb]
  \centering
  \includegraphics[height=6.5cm]{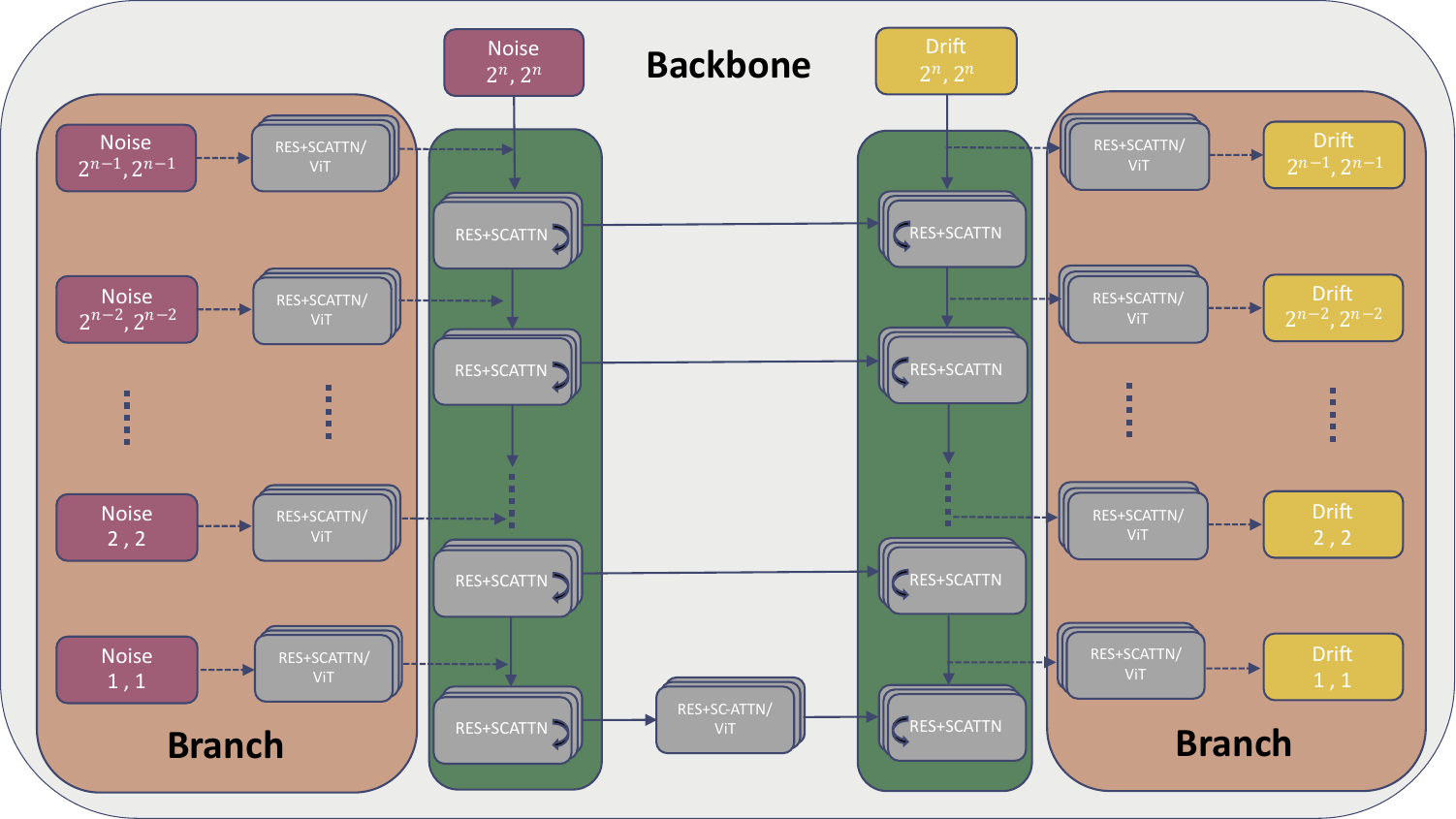}
  \caption{ The main architecture of Pyramid NCSNpp. The main architecture of Pyramid NCSNpp extends the NCSNpp with several branches, where the number of branches is determined by the number of pyramid latents adopted by the AutoEncoder. Modules connected by dotted lines represent components that are not necessarily required and are decided by the user. Deleting all modules connected by dotted lines would revert the Pyramid NCSNpp back to NCSNpp, and the framework would become the Latent Diffusion Model again with a single latent.
  }
  \label{fig:PUNET}
\end{figure}

\textbf{On UNet} Building upon NCSNplusplus proposed in \cite{song2020score}, we have implemented several modifications to the overall architecture and its components. One significant change is the incorporation of multiple branches into NCSNpp, where each branch receives distinct noise sources and generates multiple drifts for the ODE process. In our terminology, the vanilla UNet within NCSNpp is denoted as the "Backbone," while the additional components are referred to as "Branches" (See Fig.\ref{fig:PUNET}). Additionally, we have adapted the skip-connection mechanism utilized in the original NCSNpp for the branches. Specifically, we employ input-skip during down-sampling and Res-Skip during up-sampling. This involves bypassing the inputs from the branches into the backbone during down-sampling and, conversely, bypassing the features from the backbone to the outputs of the branches during up-sampling.

Unlike NCSNpp, which employs ResBlock from \cite{biggan} and Spatial attention to construct the basic module, we substitute the Spatial attention with our proposed Spatial-Channel attention. This modification offers a self-adaptation mechanism between spatial attention (Where) and channel attention (What).

\textbf{Concept Aliasing} \cite{hoogeboom2023simple} found that dropout the network only at low resolution features improve the performance of generative network. We also found that an explicit separation of phase during the process of image formation (see \ref{fig:resolutiondiff}).  Based on these two results, we assume that over-fitting for generative model is \textbf{inversely related} to level of concept abstraction in neural network, we named such assumption \textbf{Concept Aliasing}. 

Formally, \textbf{Concept Aliasing} refer to phenomenon that neural network mixing similar features with close semantic connotations when the overall constituents of image is building gradually from highly abstracted level. We assume that there exist three level of abstraction during image formation which is global concept formation, local concept formation and details refinement and concept aliasing mainly appears in the first two level in generative neural network, while in the stage of details refinement such aliasing is not significant or imperceptible. 

In first stage when global concept of image is being built, network forms outline of the whole picture and all semantic objects is roughly included in high abstraction level. For example, for " bustling churches" (see Fig.\ref{fig:resolutiondiff}, row 1,column 4) in first stage, concept of "People", "Ground", "Church" and "Sky" are generated. In this stage, over-fitting is most likely to occur because these concepts is built upon no condition (i.e. upon noise) and also it is unforced to mix concept in high abstraction level because their reference of concept is wide.(i.e. we know "people is wearing a clothes", but probably network would alias the concept of "Polo shirt" and "T-shirt" together incurring artifact "T-Polo-Shirt".)

In the second stage when local concept of image is being perceived (Fig.\ref{fig:resolutiondiff}, row 1, column 3) based on first stage, network further develop the subordinate local concept of image (e.g. "Hand of people", "Windows of the churches", "Flagpole on the ground"). In this stage, over-fitting is frequent to occur while problem is not so serious as in the first stage because features at this level is conditioning on the output of first stage and the referential scope of concepts in this stage is narrower.(e.g. Windows is part of churches, we worry that two windows is stacked on the same place, possibly with different style).

In this third stage, global and local concept is formed and clear, network further refine the details of image (e.g. texture, highlight, shadow, edge..., see Fig.\ref{fig:resolutiondiff} row 1, column 1 and 2.). In this stage, over-fitting is most unlikely to happen because both global and local concept of the image is built and network works on refining the details, even if the image is already aliased in the previous stage.

We propose decreasing dropout rate where a large dropout rate (p=0.35 typically) at features with lowest resolution and linearly decrease the dropout rate as resolution increase. We adopt this strategy because dropout at lower resolution gain more benefit on generalizability than the instability it introduces especially when network trying to fitting high frequency details such as edge under the effect of Concept Aliasing.

\section{Experiment}
\label{sec:Exper}
\subsection{Dataset}
Pyramid Diffusion could scaling up to generate really high-resolution image. For prove of concept, dataset of high resolution image is required, especially dataset with image larger than 2K resolution. However, currently dataset for image generative researches \cite{lsun}\cite{celebahq}\cite{stylegan}\cite{cifar}, to our best knowledge, are mainly concentrate on image equal or below 1024*1024 pixel. Also, dataset of image above 1024*1024 pixels which is usually used for research of image super-resolution \cite{div1}\cite{div2}\cite{div8k}\cite{ff2k} usually contain only few images where \cite{ff2k} contains 2000 images but many of them have pixels below 2048*1024. To fill this gaps on image synthesis on 2K resolution, we collect two image dataset named SCAPES2K and PEOPLE2K. SCAPES2K contains about 13000 images with the theme of natural scenes and all images have pixel above 2048*1024 pixels. For SCAPES2K, we collected about 30000 images from Flickr, then we pick up all the images with width larger than 2048 pixels and height larger than 1024 pixels and with width larger than height. Finally, we get 12789 images of natural scenes with 2K resolution and most of the image has aspect ratio close to 2. For PEOPLE2K, we collected about 20000 images from Unsplash which contain human portraits including body and face, then we pick up all the images with both height and width larger than 2048 and then we crop the image with its height or width equaling to the smallest one. Specifically, if height is 3200 and width is 2400 pixels, we crop the image with width equaling to 2400 pixels and we cut the image along height on both side with (3200 -2400)/2 = 400 pixels, resulting image with height equaling to 2400 pixels. After processing and selection, PEOPLE2K contain about 4846 images. Beside these two datasets, we also train our model on datasets with lower resolution, they are LSUN-Bedrooms, LSUN-Churches and CelebaHQ (1024 * 1024)

\subsection{Unconditioned Image Generation}
To show whether Pyramid Diffusion model could scaling up to generate images with ultra-high-resolution, we test our model on SCAPES2K and PEOPLE2K and three high-resolution datasets. The methods are evaluated in variety and quality by Fréchet inception distance (FID). For LSUN-Bedrooms we sample 200K images for training \cite{wang2022diffusiongan} and use all images for training for other datasets. We evaluate fid with 50K generated image for LSUN-Bedroom and LSUN-Churches, 30K generated image for Celeba-1024 and 10K for SCAPES2K and 5k for PEOPLE2K on all images in training set.

\textbf{Experiment Setting} We train LSUN-Bedrooms and LSUN-Churches on single RTX 4090 with 24GB VRAM and CelebaHQ, SCAPES2K and PEOPLE2K on single A100 with 40GB VRAM. We also rescale the latent such that rescaled latent has unit standard deviation, we estimate the standard deviation with exponential moving average for 100 iterations in the very beginning of training and skip the rescaling on latent with standard deviation smaller than 1. When training AutoEncoder on SCAPES2K, PEOPLE2K,  we use batch size 1 and accumulate the gradient to get effective batch size 4. More details on architectures and hyperparameters can be found at the appendix.

\textbf{Sampling} We use black box ode solver 'RK45' and plain euler 
 with 200 steps for sampling. Also, to relieve the under-sample problem at the boundary and other potential boundary problems, we slightly enlarge the sampling range where $t\in[eps,1 + eps]$ during training but with maximum of t clamped to 1, while during sampling we evolve with $t\in(eps,1]$

\begin{table}[ht]

\resizebox{0.5\linewidth}{!}{
\centering

\begin{minipage}[b]{0.5\linewidth}
    \centering
    \scalebox{0.9}{
    \begin{tabular}{c|c|ccc}
    \toprule
    Dataset & Method & FID $\downarrow$\\
    \midrule
    \multirow{4}{*}{LSUN-Churches $256 \times 256$} & Patch Diffusion\cite{wang2023patch} & 2.66 \\
                                                 & Diffusion GAN\cite{kumari2022visionaid} & 1.85\\
                                                 & Projected GAN\cite{projectedgan} & 1.59 \\
                                                 & Pyramid Diffuson & - \\
    \midrule
    \multirow{5}{*}{CelebA-HQ $1024 \times 1024$} 
                                                 & PG-SWGAN\cite{wu2019sliced} & 5.5	 \\
                                                 & StyleGan\cite{stylegan} & 5.06  \\
                                                 & StyleSwin\cite{zhang2022styleswin} & 	4.43  \\
                                                 & Pyramid Diffuson  & -  \\
    \bottomrule
    \end{tabular}}
\end{minipage}}
\resizebox{0.5\linewidth}{!}{
\centering
\begin{minipage}[b]{0.5\linewidth}
\centering
\scalebox{0.87}{
    \begin{tabular}{c|c|ccc}
    \toprule
    Dataset & Method & FID $\downarrow$ \\
    \midrule
    \multirow{4}{*}{LSUN-Bedrooms $256 \times 256$} & TDPM+\cite{zheng2023truncated} & 1.88 \\
                                                 & Projected GAN\cite{projectedgan} & 1.52\\
                                                 & Diffusion-GAN\cite{wang2022diffusiongan} & 1.43 \\
                                                 & Pyramid Diffusion  & -  \\
    \midrule
    \multirow{2}{*}{SCAPES2K $1024 \times 2048$} &  &  \\
                                                 & Pyramid Diffusion  &   \\
    \midrule
    \multirow{3}{*}{PEOPLE2K $2048 \times 2048$} & &   \\
                                                 & Pyramid Diffusion  & -  \\
    \bottomrule
    \end{tabular}}
    
\end{minipage}}
    \caption{FID Score for unconditional image synthesis using RK45 Sampler. $\star:$ N-s refer to the average steps RK45 sampler used for sampling}
    \label{table:unconditional}
\end{table}
\textbf{Results} As shown in Table.\ref{table:unconditional},

\subsection{Visualization of Pyramid Latent}
\label{subsec:VisPyramid}
\begin{figure}[htb]
  \centering
  \includegraphics[height=7cm]{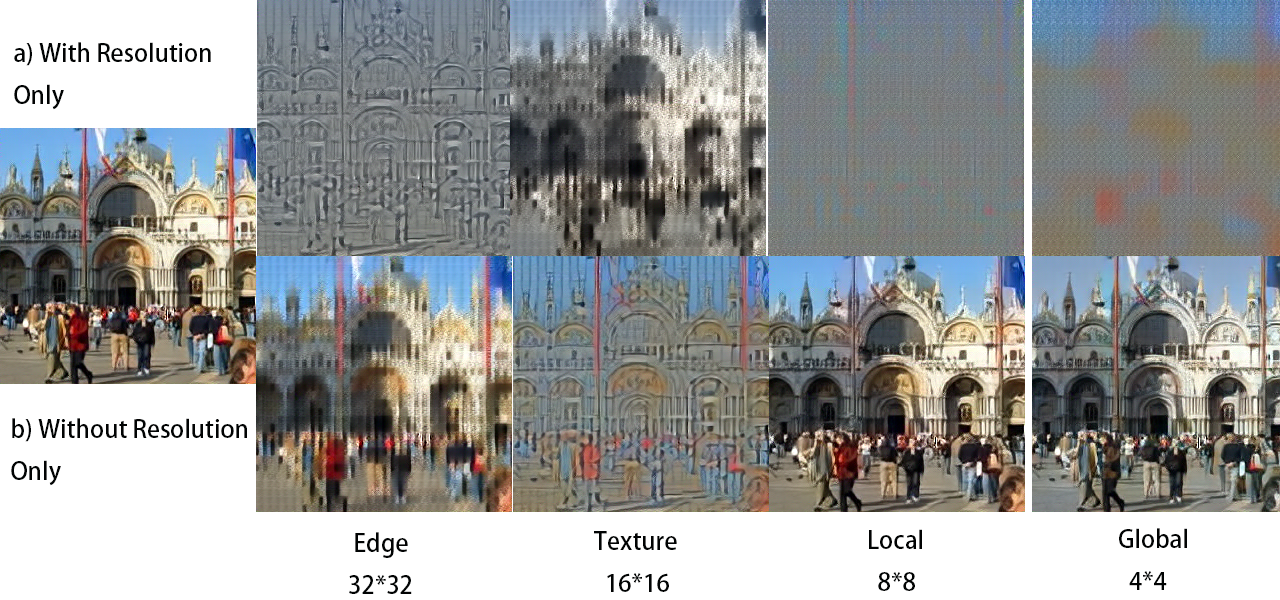}
  \caption{The visualization depicts how latents in different layers affect the final image. The image on the leftmost side is constructed using all latents. The first row displays images built using only one latent with a specific resolution. Meanwhile, the second row showcases images reconstructed using all latents except one with a specific resolution. The specific resolution is listed at the bottom of each image. By observing these visualizations, we can discern that the formation of the image is subdivided into several sub-tasks. Progressively, the concept and details of the image become clearer as the reconstruction process proceeds.}
  \label{fig:resolutiondiff}
\end{figure}

 We visualize pyramid latent(See Fig.\ref{fig:resolutiondiff}). We design a control experiment where we control the latent in different resolution level. For each pyramid latent, we exclude only or include only the specific latent. Fig.\ref{fig:resolutiondiff} show the results of experiment. The single image at the leftmost is reconstructed with complete pyramid latent. Images in the row 1 are reconstructed with only \textbf{one} latent with specific resolution, while images in the row 2 are reconstructed without only \textbf{one} latent with specific resolution
 
 Results from control experiment on pyramid latent show that latent with different resolution undertake different part of the image. In the bottom layers of UNet where 4*4 latent is decoded into image, concept and global color is formed; without 4*4 latent, image become unsaturated. With 8*8 latent, the local concept and color is reinforced into image; without 8*8 latent local details of color will loss. Later, when decoder receive 16*16 latent and further build image with texture conditioned on previous layers, the concept of the image is clear with only 16*16 latent; Without 16*16 latent, the image undergo problem of Texture Loss. With 32*32 latent, decoder further add high-frequency information into image, such as edge; without 32*32 latent, the boundary inside or outside the object becomes faint.

\section{Conclusion} 

\textbf{Conclusion} In this passage, we proposed a pyramid latent representation and several modifications on basic component inside the neural network for high-resolution image synthesis. Pyramid Diffusion Model achieves to generate extremely high-resolution image with 2K resolution for the first time to the best of our knowledge.

\section*{Acknowledgments}
This research was self-funded. I am deeply grateful for the encouragement and support from my friends and family throughout this endeavor.

\bibliographystyle{splncs04}  
\bibliography{references}  

\begin{thebibliography}{10}
\providecommand{\url}[1]{\texttt{#1}}
\providecommand{\urlprefix}{URL }
\providecommand{\doi}[1]{https://doi.org/#1}

\bibitem{div2}
Agustsson, E., Timofte, R.: Ntire 2017 challenge on single image super-resolution: Dataset and study. In: The IEEE Conference on Computer Vision and Pattern Recognition (CVPR) Workshops (July 2017)

\bibitem{biggan}
Andrew~Brock, J.D., Simonyan, K.: Large scale gan training for high fidelity natural image synthesis. In: ICLR (2019)

\bibitem{WGAN}
Arjovsky, M., Chintala, S., Bottou, L.: Wasserstein generative adversarial networks. In: International conference on machine learning. pp. 214--223. PMLR (2017)

\bibitem{bao2023ViTLDM}
Bao, F., Nie, S., Xue, K., Cao, Y., Li, C., Su, H., Zhu, J.: All are worth words: A vit backbone for diffusion models (2023)

\bibitem{vecGanpp}
Dalva, Y., Pehlivan, H., Hatipoglu, O.I., Moran, C., Dundar, A.: Image-to-image translation with disentangled latent vectors for face editing. IEEE Transactions on Pattern Analysis and Machine Intelligence  (2023)

\bibitem{dhariwal2021diffusionbeatgan}
Dhariwal, P., Nichol, A.: Diffusion models beat gans on image synthesis. Advances in neural information processing systems  \textbf{34},  8780--8794 (2021)

\bibitem{dinh2014nice}
Dinh, L., Krueger, D., Bengio, Y.: Nice: Non-linear independent components estimation. arXiv preprint arXiv:1410.8516  (2014)

\bibitem{realNVP}
Dinh, L., Sohl-Dickstein, J., Bengio, S.: Density estimation using real nvp. arXiv preprint arXiv:1605.08803  (2016)

\bibitem{eldan2016power}
Eldan, R., Shamir, O.: The power of depth for feedforward neural networks (2016)

\bibitem{esser2020taming}
Esser, P., Rombach, R., Ommer, B.: Taming transformers for high-resolution image synthesis (2020)

\bibitem{gan2014}
Goodfellow, I., Pouget-Abadie, J., Mirza, M., Xu, B., Warde-Farley, D., Ozair, S., Courville, A., Bengio, Y.: Generative adversarial nets. Advances in neural information processing systems  \textbf{27} (2014)

\bibitem{div8k}
Gu, S., Lugmayr, A., Danelljan, M., Fritsche, M., Lamour, J., Timofte, R.: Div8k: Diverse 8k resolution image dataset. In: 2019 IEEE/CVF International Conference on Computer Vision Workshop (ICCVW). pp. 3512--3516 (2019). \doi{10.1109/ICCVW.2019.00435}

\bibitem{gulrajani2017improvedWGANResNet}
Gulrajani, I., Ahmed, F., Arjovsky, M., Dumoulin, V., Courville, A.: Improved training of wasserstein gans (2017)

\bibitem{WGAN-GP}
Gulrajani, I., Ahmed, F., Arjovsky, M., Dumoulin, V., Courville, A.C.: Improved training of wasserstein gans. Advances in neural information processing systems  \textbf{30} (2017)

\bibitem{fid}
Heusel, M., Ramsauer, H., Unterthiner, T., Nessler, B., Hochreiter, S.: Gans trained by a two time-scale update rule converge to a local nash equilibrium. Advances in neural information processing systems  \textbf{30} (2017)

\bibitem{ho2020ddpm}
Ho, J., Jain, A., Abbeel, P.: Denoising diffusion probabilistic models. Advances in neural information processing systems  \textbf{33},  6840--6851 (2020)

\bibitem{hoogeboom2023simple}
Hoogeboom, E., Heek, J., Salimans, T.: Simple diffusion: End-to-end diffusion for high resolution images (2023)

\bibitem{karnewar2020msggan}
Karnewar, A., Wang, O.: Msg-gan: Multi-scale gradients for generative adversarial networks (2020)

\bibitem{karras2018progressive}
Karras, T., Aila, T., Laine, S., Lehtinen, J.: Progressive growing of gans for improved quality, stability, and variation (2018)

\bibitem{celebahq}
Karras, T., Aila, T., Laine, S., Lehtinen, J.: Progressive growing of gans for improved quality, stability, and variation (2018)

\bibitem{karras2022elucidating}
Karras, T., Aittala, M., Aila, T., Laine, S.: Elucidating the design space of diffusion-based generative models (2022)

\bibitem{stylegan}
Karras, T., Laine, S., Aila, T.: A style-based generator architecture for generative adversarial networks. In: Proceedings of the IEEE/CVF conference on computer vision and pattern recognition. pp. 4401--4410 (2019)

\bibitem{styleganplus}
Karras, T., Laine, S., Aittala, M., Hellsten, J., Lehtinen, J., Aila, T.: Analyzing and improving the image quality of stylegan. pp. 8107--8116 (06 2020). \doi{10.1109/CVPR42600.2020.00813}

\bibitem{karras2020analyzing}
Karras, T., Laine, S., Aittala, M., Hellsten, J., Lehtinen, J., Aila, T.: Analyzing and improving the image quality of stylegan (2020)

\bibitem{VAE}
Kingma, D.P., Welling, M.: Auto-encoding variational bayes. arXiv preprint arXiv:1312.6114  (2013)

\bibitem{cifar}
Krizhevsky, A., Hinton, G., et~al.: Learning multiple layers of features from tiny images  (2009)

\bibitem{kumari2022visionaid}
Kumari, N., Zhang, R., Shechtman, E., Zhu, J.Y.: Ensembling off-the-shelf models for gan training (2022)

\bibitem{precision}
Kynk{\"a}{\"a}nniemi, T., Karras, T., Laine, S., Lehtinen, J., Aila, T.: Improved precision and recall metric for assessing generative models. Advances in Neural Information Processing Systems  \textbf{32} (2019)

\bibitem{li2022supervisedGan}
Li, N., Plummer, B.A.: Supervised attribute information removal and reconstruction for image manipulation (2022)

\bibitem{ff2k}
Lim, B., Son, S., Kim, H., Nah, S., Lee, K.M.: Enhanced deep residual networks for single image super-resolution. In: The IEEE Conference on Computer Vision and Pattern Recognition (CVPR) Workshops (July 2017)

\bibitem{ling2021editgan}
Ling, H., Kreis, K., Li, D., Kim, S.W., Torralba, A., Fidler, S.: Editgan: High-precision semantic image editing. Advances in Neural Information Processing Systems  \textbf{34},  16331--16345 (2021)

\bibitem{liu2022rectflow}
Liu, X., Gong, C., Liu, Q.: Flow straight and fast: Learning to generate and transfer data with rectified flow (2022)

\bibitem{lu2022dpmsolver}
Lu, C., Zhou, Y., Bao, F., Chen, J., Li, C., Zhu, J.: Dpm-solver: A fast ode solver for diffusion probabilistic model sampling in around 10 steps (2022)

\bibitem{unstableGan}
Metz, L., Poole, B., Pfau, D., Sohl-Dickstein, J.: Unrolled generative adversarial networks. arXiv preprint arXiv:1611.02163  (2016)

\bibitem{miyato2018spectral}
Miyato, T., Kataoka, T., Koyama, M., Yoshida, Y.: Spectral normalization for generative adversarial networks (2018)

\bibitem{sparse-representation-gen}
Nash, C., Menick, J., Dieleman, S., Battaglia, P.W.: Generating images with sparse representations. arXiv preprint arXiv:2103.03841  (2021)

\bibitem{ddim}
Nichol, A., Dhariwal, P.: Improved denoising diffusion probabilistic models (2021)

\bibitem{nichol2022glide}
Nichol, A., Dhariwal, P., Ramesh, A., Shyam, P., Mishkin, P., McGrew, B., Sutskever, I., Chen, M.: Glide: Towards photorealistic image generation and editing with text-guided diffusion models (2022)

\bibitem{IDDPM}
Nichol, A.Q., Dhariwal, P.: Improved denoising diffusion probabilistic models. In: International Conference on Machine Learning. pp. 8162--8171. PMLR (2021)

\bibitem{pixelcnn}
Van~den Oord, A., Kalchbrenner, N., Espeholt, L., Vinyals, O., Graves, A., et~al.: Conditional image generation with pixelcnn decoders. Advances in neural information processing systems  \textbf{29} (2016)

\bibitem{vqvae}
van~den Oord, A., Vinyals, O., Kavukcuoglu, K.: Neural discrete representation learning (2018)

\bibitem{pan2023drag}
Pan, X., Tewari, A., Leimk{\"u}hler, T., Liu, L., Meka, A., Theobalt, C.: Drag your gan: Interactive point-based manipulation on the generative image manifold. In: ACM SIGGRAPH 2023 Conference Proceedings. pp. 1--11 (2023)

\bibitem{peebles2023scalable}
Peebles, W., Xie, S.: Scalable diffusion models with transformers (2023)

\bibitem{pernias2023wuerstchen}
Pernias, P., Rampas, D., Richter, M.L., Pal, C.J., Aubreville, M.: Wuerstchen: An efficient architecture for large-scale text-to-image diffusion models (2023)

\bibitem{podell2023sdxl}
Podell, D., English, Z., Lacey, K., Blattmann, A., Dockhorn, T., Müller, J., Penna, J., Rombach, R.: Sdxl: Improving latent diffusion models for high-resolution image synthesis (2023)

\bibitem{ramesh2022hierarchicalt2i}
Ramesh, A., Dhariwal, P., Nichol, A., Chu, C., Chen, M.: Hierarchical text-conditional image generation with clip latents (2022)

\bibitem{ramesh2021zeroshot}
Ramesh, A., Pavlov, M., Goh, G., Gray, S., Voss, C., Radford, A., Chen, M., Sutskever, I.: Zero-shot text-to-image generation (2021)

\bibitem{vqvae2}
Razavi, A., Van~den Oord, A., Vinyals, O.: Generating diverse high-fidelity images with vq-vae-2. Advances in neural information processing systems  \textbf{32} (2019)

\bibitem{ldm}
Rombach, R., Blattmann, A., Lorenz, D., Esser, P., Ommer, B.: High-resolution image synthesis with latent diffusion models (2021)

\bibitem{ronneberger2015unet}
Ronneberger, O., Fischer, P., Brox, T.: U-net: Convolutional networks for biomedical image segmentation (2015)

\bibitem{ISScore}
Salimans, T., Goodfellow, I., Zaremba, W., Cheung, V., Radford, A., Chen, X.: Improved techniques for training gans. Advances in neural information processing systems  \textbf{29} (2016)

\bibitem{projectedgan}
Sauer, A., Chitta, K., M{\"{u}}ller, J., Geiger, A.: Projected gans converge faster. In: Advances in Neural Information Processing Systems (NeurIPS) (2021)

\bibitem{styleganXL}
Sauer, A., Schwarz, K., Geiger, A.: Stylegan-xl: Scaling stylegan to large diverse datasets. vol. abs/2201.00273 (2022), \url{https://arxiv.org/abs/2201.00273}

\bibitem{sinha2021d2c}
Sinha, A., Song, J., Meng, C., Ermon, S.: D2c: Diffusion-denoising models for few-shot conditional generation (2021)

\bibitem{song2023consistencymodel}
Song, Y., Dhariwal, P., Chen, M., Sutskever, I.: Consistency models (2023)

\bibitem{song2020score}
Song, Y., Sohl-Dickstein, J., Kingma, D.P., Kumar, A., Ermon, S., Poole, B.: Score-based generative modeling through stochastic differential equations. arXiv preprint arXiv:2011.13456  (2020)

\bibitem{fvae}
Su, J., Wu, G.: f-vaes: Improve vaes with conditional flows. arXiv preprint arXiv:1809.05861  (2018)

\bibitem{div1}
Timofte, R., Agustsson, E., Van~Gool, L., Yang, M.H., Zhang, L., Lim, B., et~al.: Ntire 2017 challenge on single image super-resolution: Methods and results. In: The IEEE Conference on Computer Vision and Pattern Recognition (CVPR) Workshops (July 2017)

\bibitem{pixelrnn}
Van Den~Oord, A., Kalchbrenner, N., Kavukcuoglu, K.: Pixel recurrent neural networks. In: International conference on machine learning. pp. 1747--1756. PMLR (2016)

\bibitem{stylenat}
Walton, S., Hassani, A., Xu, X., Wang, Z., Shi, H.: Stylenat: Giving each head a new perspective  (2022), \url{https://arxiv.org/abs/2211.05770}

\bibitem{wang2023patch}
Wang, Z., Jiang, Y., Zheng, H., Wang, P., He, P., Wang, Z., Chen, W., Zhou, M.: Patch diffusion: Faster and more data-efficient training of diffusion models (2023)

\bibitem{wang2022diffusiongan}
Wang, Z., Zheng, H., He, P., Chen, W., Zhou, M.: Diffusion-gan: Training gans with diffusion. arXiv preprint arXiv:2206.02262  (2022)

\bibitem{wu2019sliced}
Wu, J., Huang, Z., Acharya, D., Li, W., Thoma, J., Paudel, D.P., Gool, L.V.: Sliced wasserstein generative models (2019)

\bibitem{xing2023diffusionvideosurvey}
Xing, Z., Feng, Q., Chen, H., Dai, Q., Hu, H., Xu, H., Wu, Z., Jiang, Y.G.: A survey on video diffusion models (2023)

\bibitem{2019propagation}
Xu, D., Lee, M.L., Hsu, W.: Propagation mechanism for deep and wide neural networks. In: 2019 IEEE/CVF Conference on Computer Vision and Pattern Recognition (CVPR). pp. 9212--9220 (2019). \doi{10.1109/CVPR.2019.00944}

\bibitem{yoshida2017spectral}
Yoshida, Y., Miyato, T.: Spectral norm regularization for improving the generalizability of deep learning (2017)

\bibitem{lsun}
Yu, F., Zhang, Y., Song, S., Seff, A., Xiao, J.: Lsun: Construction of a large-scale image dataset using deep learning with humans in the loop. CoRR  \textbf{abs/1506.03365} (2015), \url{http://dblp.uni-trier.de/db/journals/corr/corr1506.html#YuZSSX15}

\bibitem{zagoruyko2017wide}
Zagoruyko, S., Komodakis, N.: Wide residual networks (2017)

\bibitem{zhang2022styleswin}
Zhang, B., Gu, S., Zhang, B., Bao, J., Chen, D., Wen, F., Wang, Y., Guo, B.: Styleswin: Transformer-based gan for high-resolution image generation (2022)

\bibitem{zhao2022egsde}
Zhao, M., Bao, F., Li, C., Zhu, J.: Egsde: Unpaired image-to-image translation via energy-guided stochastic differential equations (2022)

\bibitem{zheng2023truncated}
Zheng, H., He, P., Chen, W., Zhou, M.: Truncated diffusion probabilistic models and diffusion-based adversarial auto-encoders (2023)

\end{thebibliography}
%
%
\newpage
\begin{center}
\textbf{Appendix}
\end{center}

\appendix

\section{Details on Implementation}
\section{Unconditional Samples}
\subsection{More samples on SCAPES2K}
1 \\
\begin{figure}[htbp]	
\vspace{-10mm} 
\centering	
\begin{minipage}[t]{1\textwidth}
	\centering
  \includegraphics[height =0.5\linewidth]{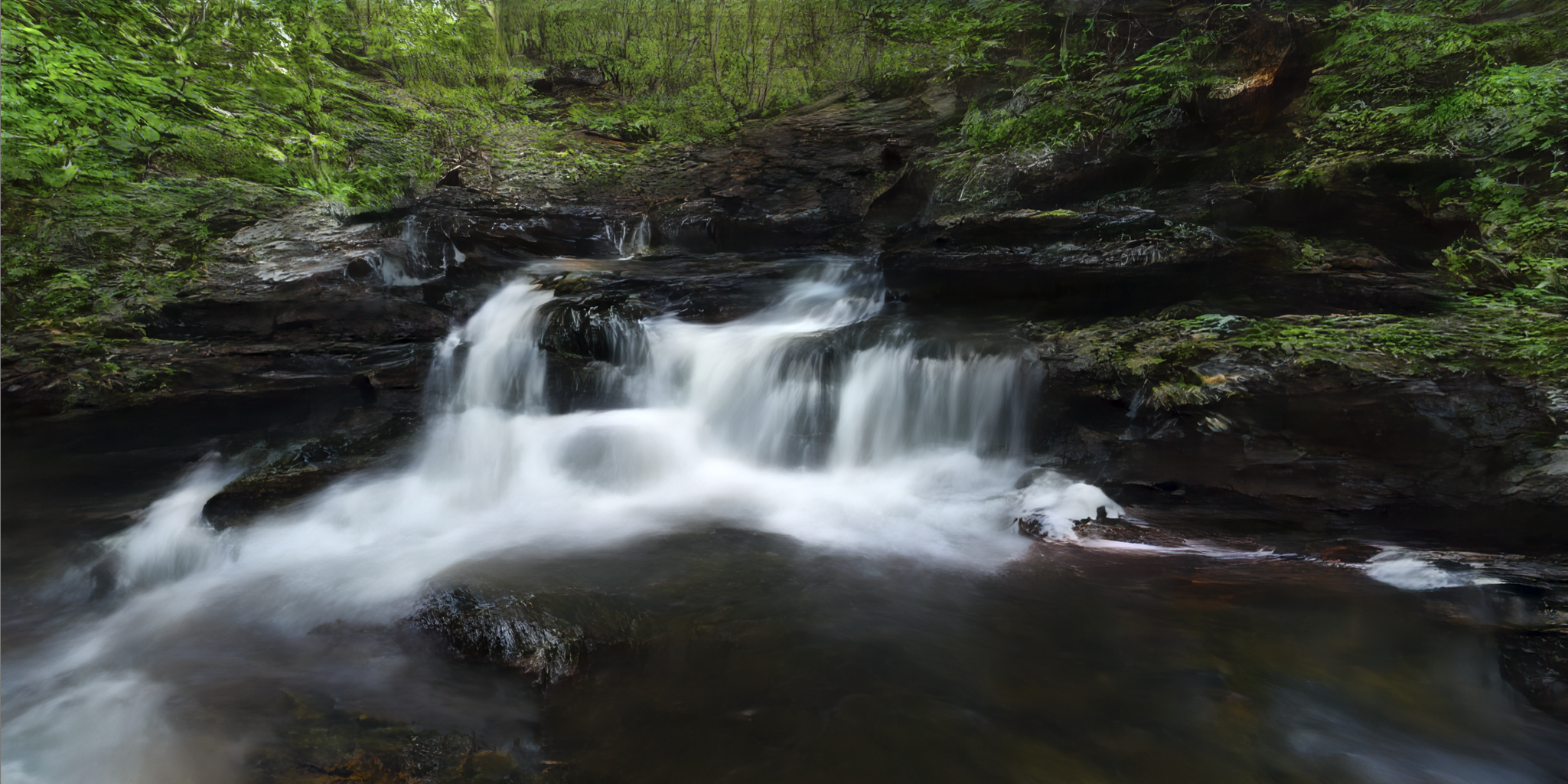}
		\label{pretrain(a)}
		\vspace{-10mm} 
\end{minipage}
\begin{minipage}[t]{1\textwidth}
	\centering
  \includegraphics[height =0.5\linewidth]{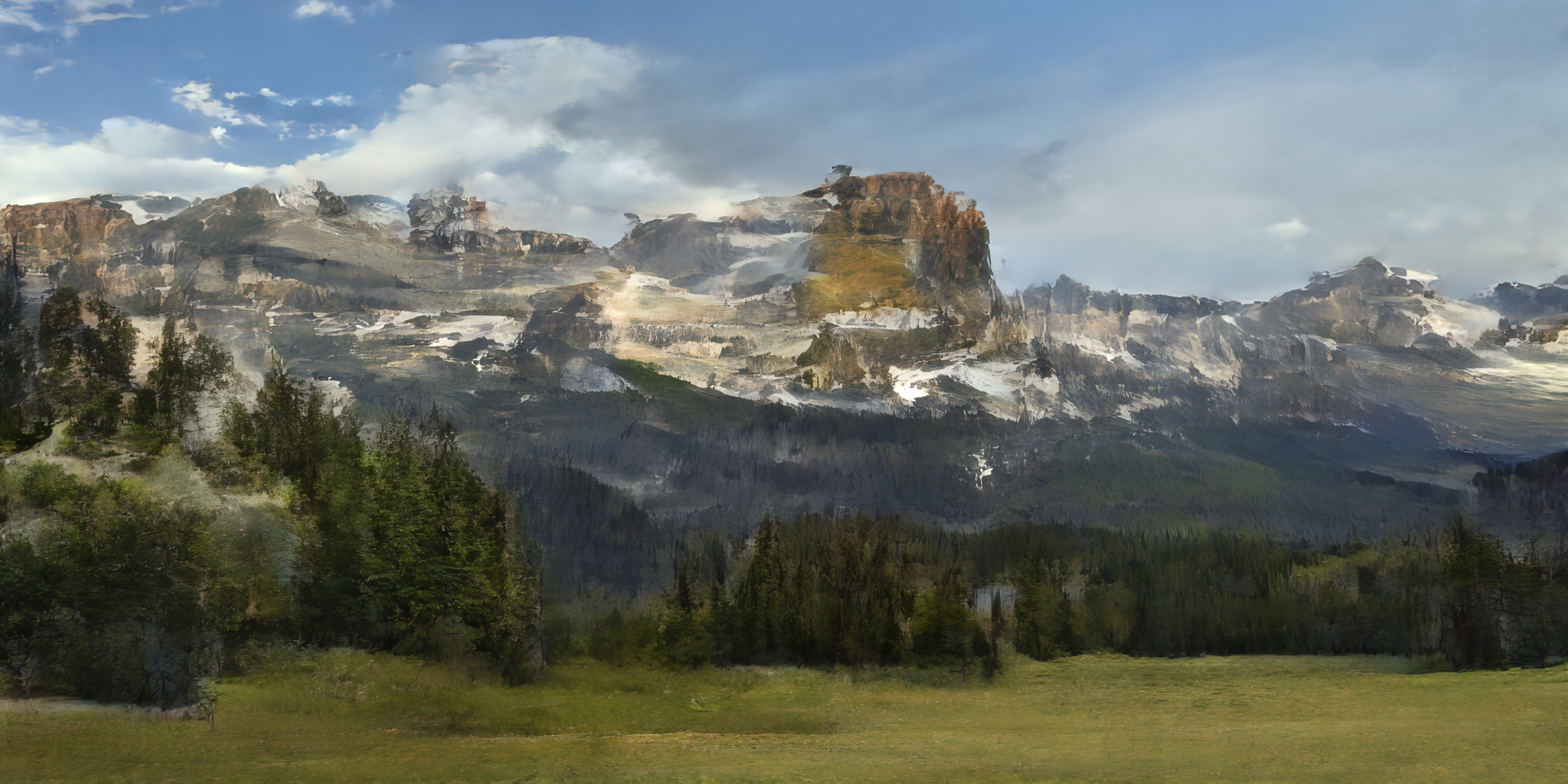}
\end{minipage}

  %

  %
\end{figure}

\begin{figure}[hbp]	
\vspace{-10mm} 
\centering	
\begin{minipage}[t]{1\textwidth}
	\centering
  \includegraphics[height =0.49\linewidth]{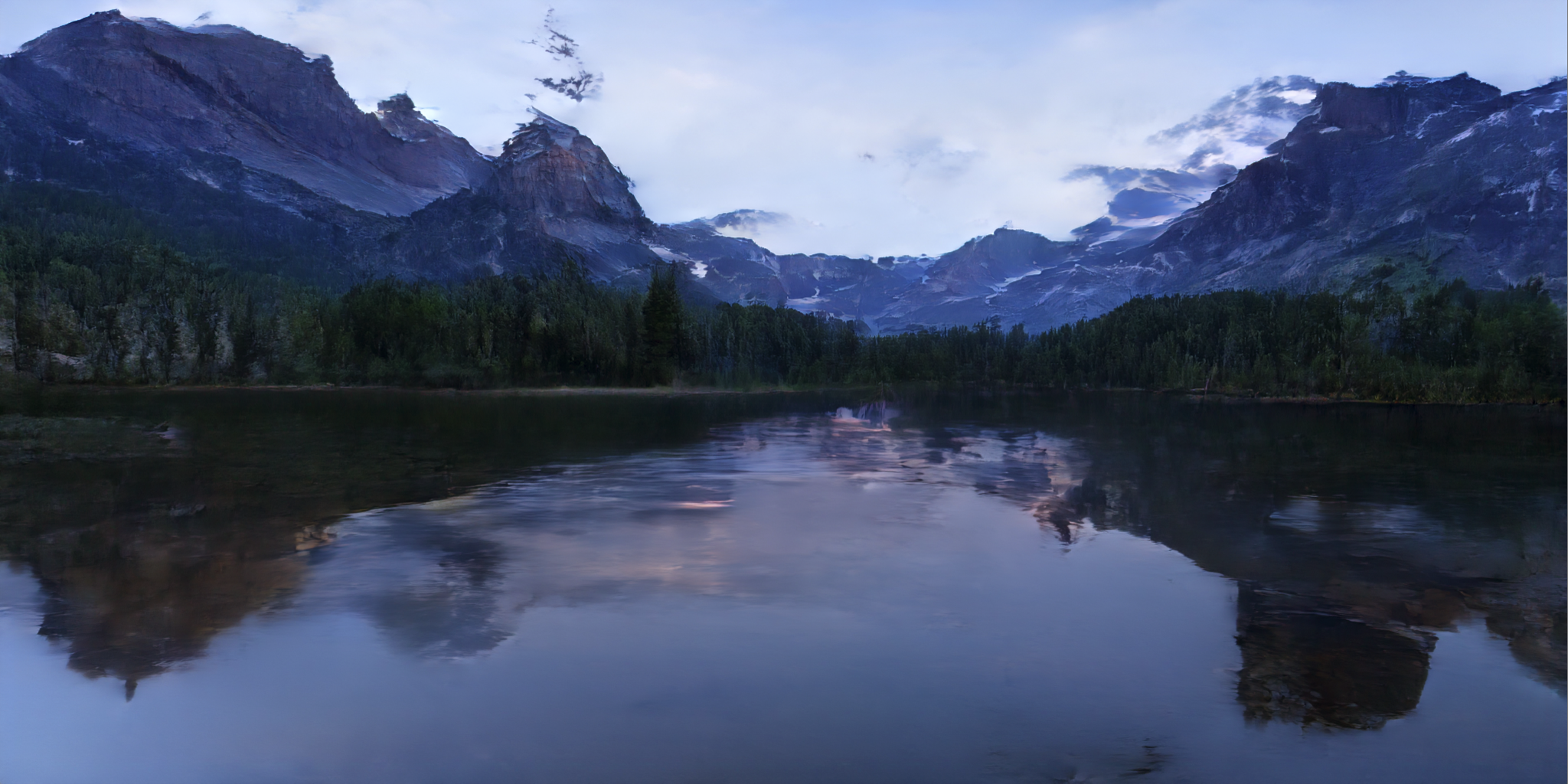}
		\label{pretrain(a)}
		\vspace{-10mm} 
\end{minipage}
\begin{minipage}[t]{1\textwidth}
	\centering
  \includegraphics[height =0.49\linewidth]{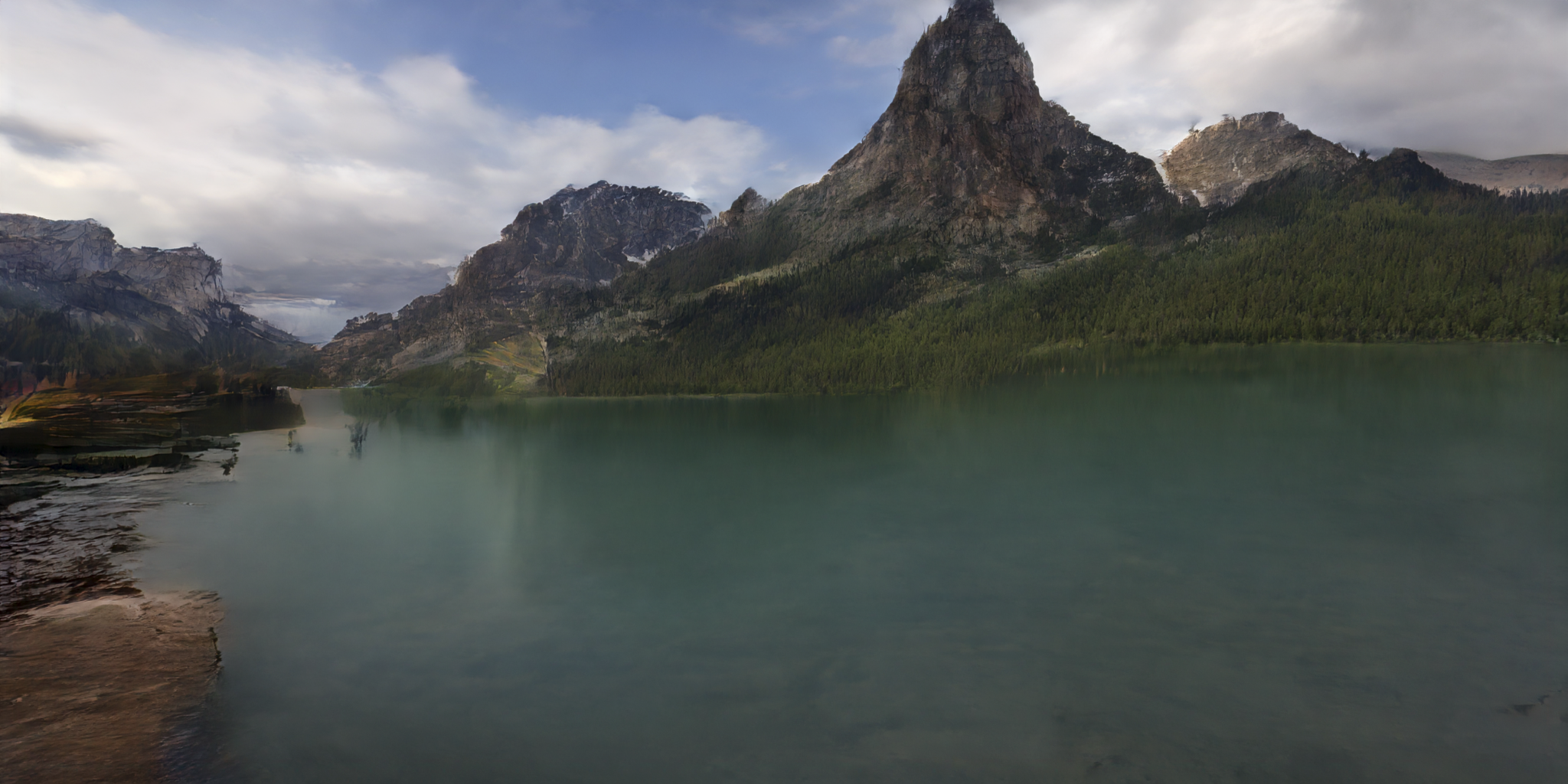}
  
\end{minipage}
\begin{minipage}[t]{1\textwidth}
	\centering
  \includegraphics[height =0.49\linewidth]{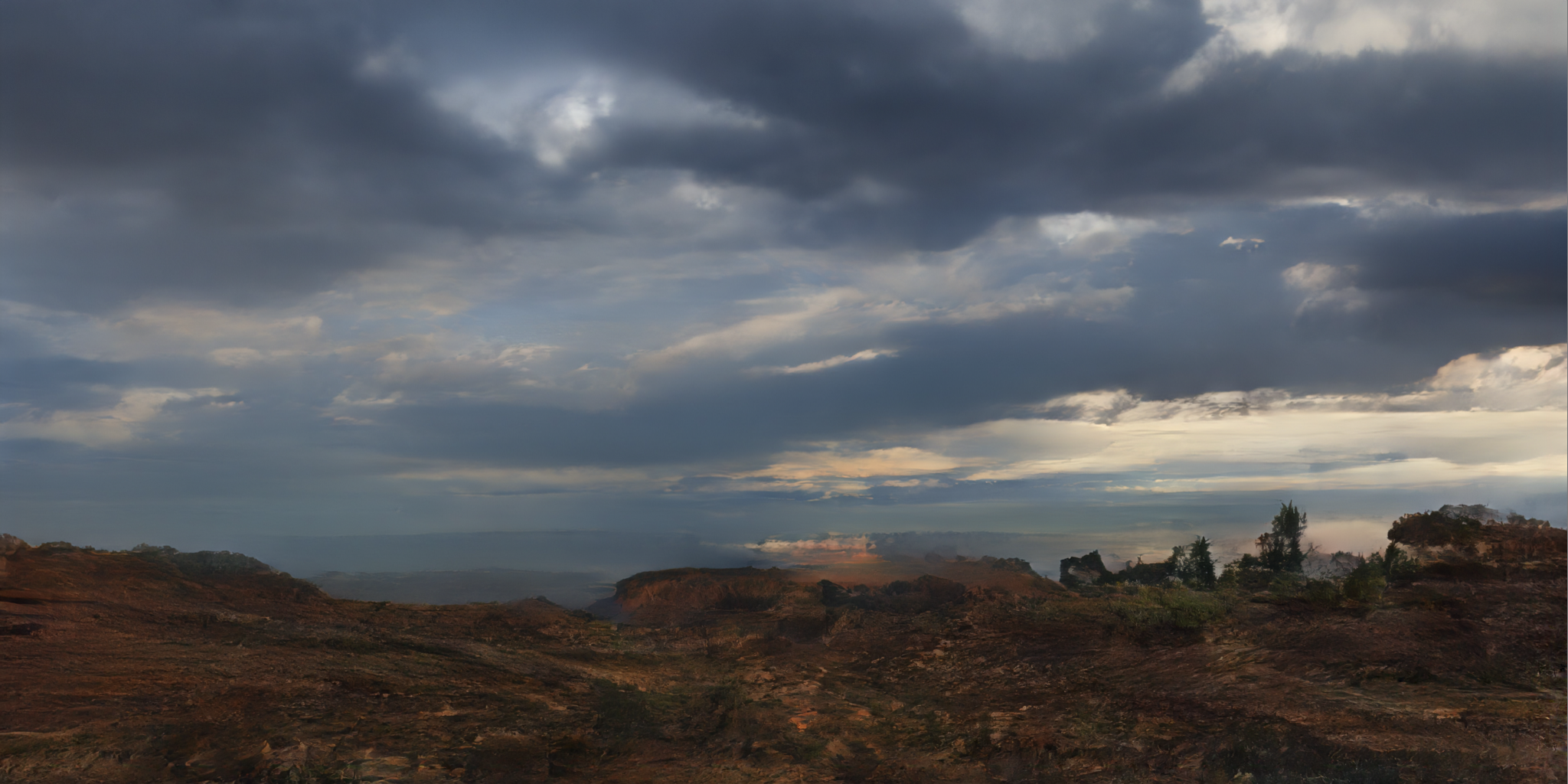}
  
\end{minipage}
\end{figure}

\begin{figure}[hbp]	
\vspace{-10mm} 
\centering	
\begin{minipage}[t]{1\textwidth}
	\centering
  \includegraphics[height =0.49\linewidth]{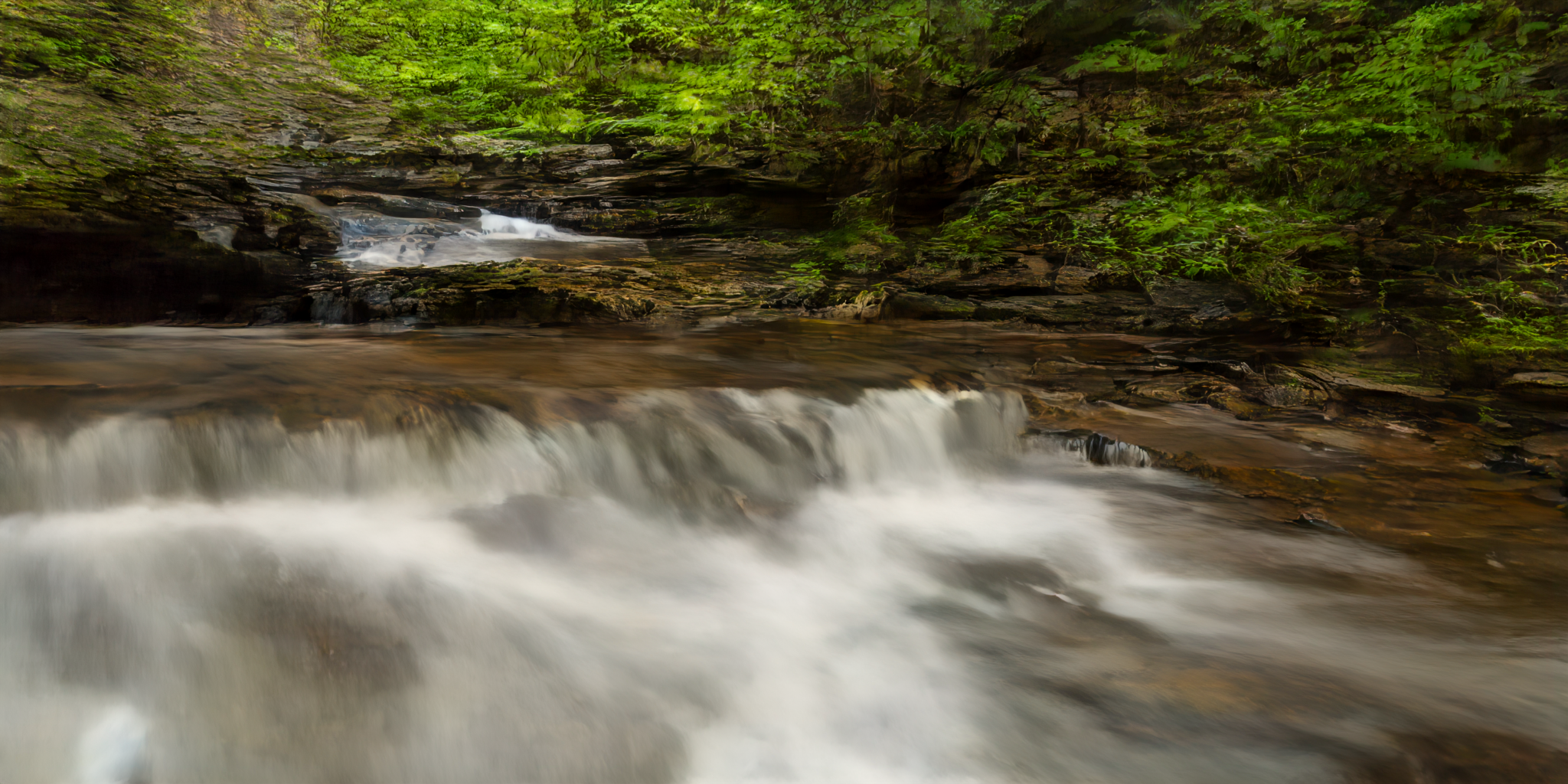}
		\label{pretrain(a)}
		\vspace{-10mm} 
\end{minipage}
\begin{minipage}[t]{1\textwidth}
	\centering
  \includegraphics[height =0.49\linewidth]{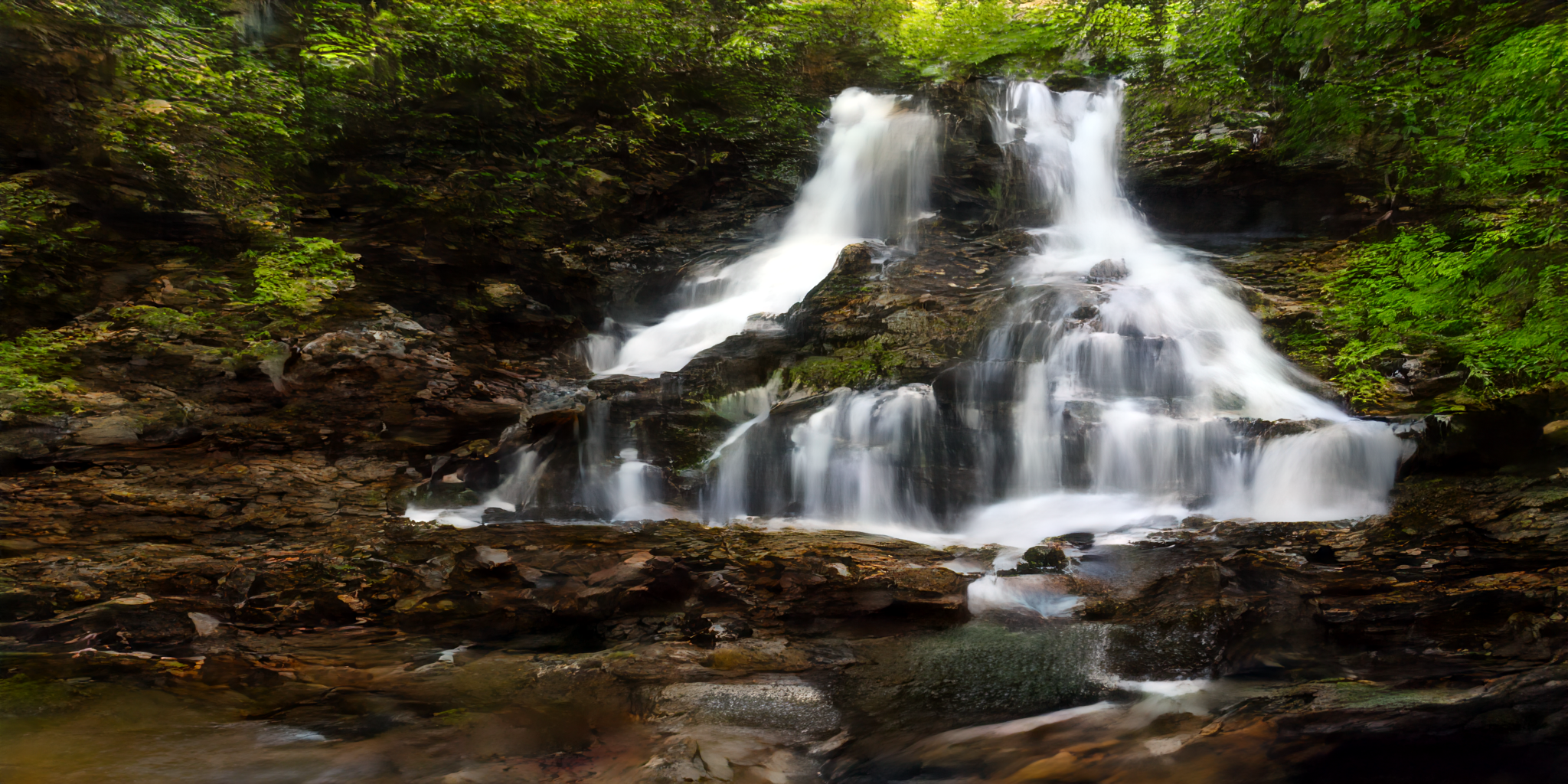}
  
\end{minipage}
\begin{minipage}[t]{1\textwidth}
	\centering
  \includegraphics[height =0.49\linewidth]{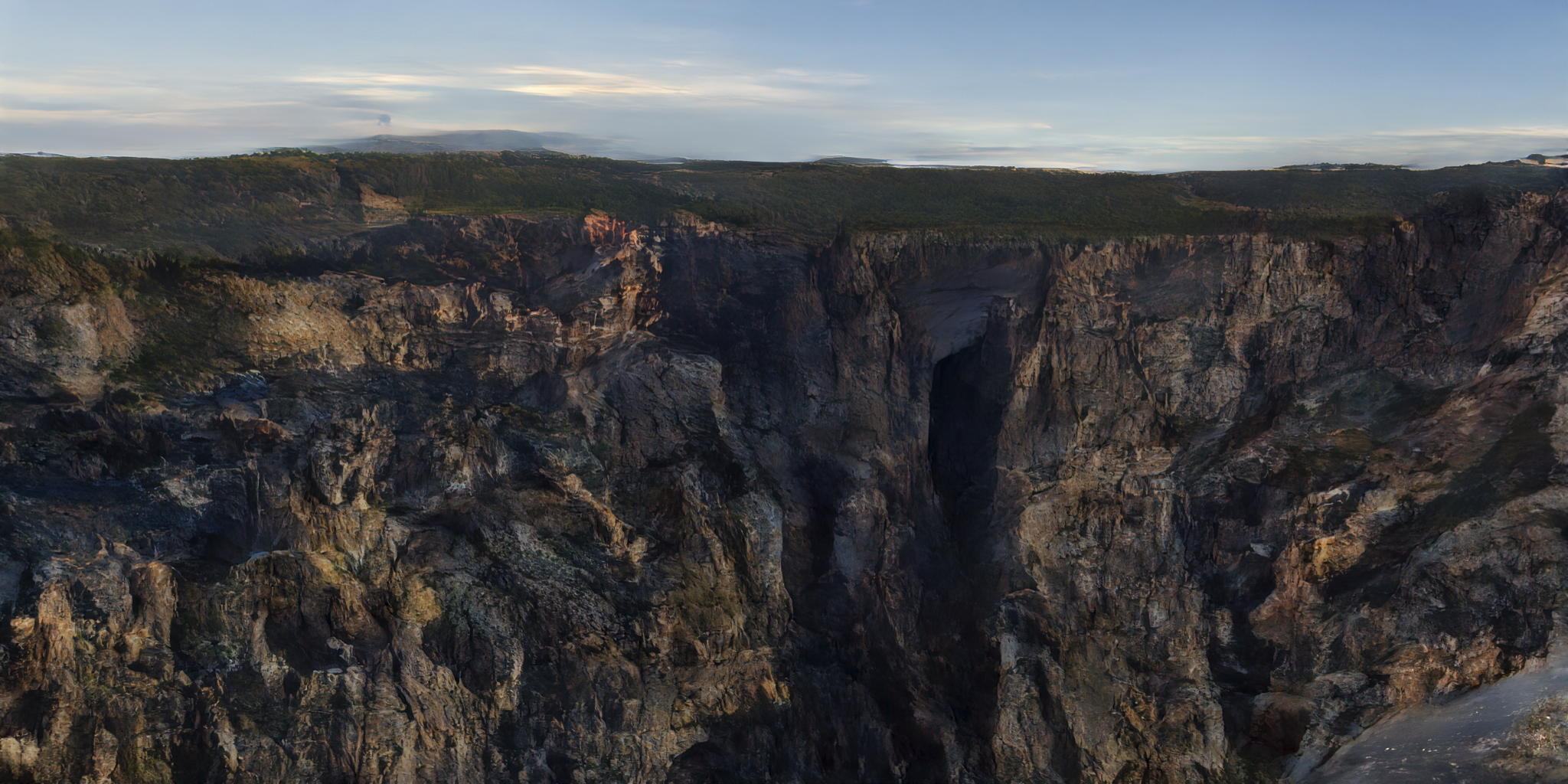}
  \caption{Generated Image, sampling with RK45, intermediate results}
\end{minipage}

\end{figure}
\newpage

\subsection{More samples on People2K}
\label{samplepeople}
\begin{figure}[htbp]
\begin{minipage}{\textwidth}
\includegraphics[width=\textwidth,height=\textwidth]{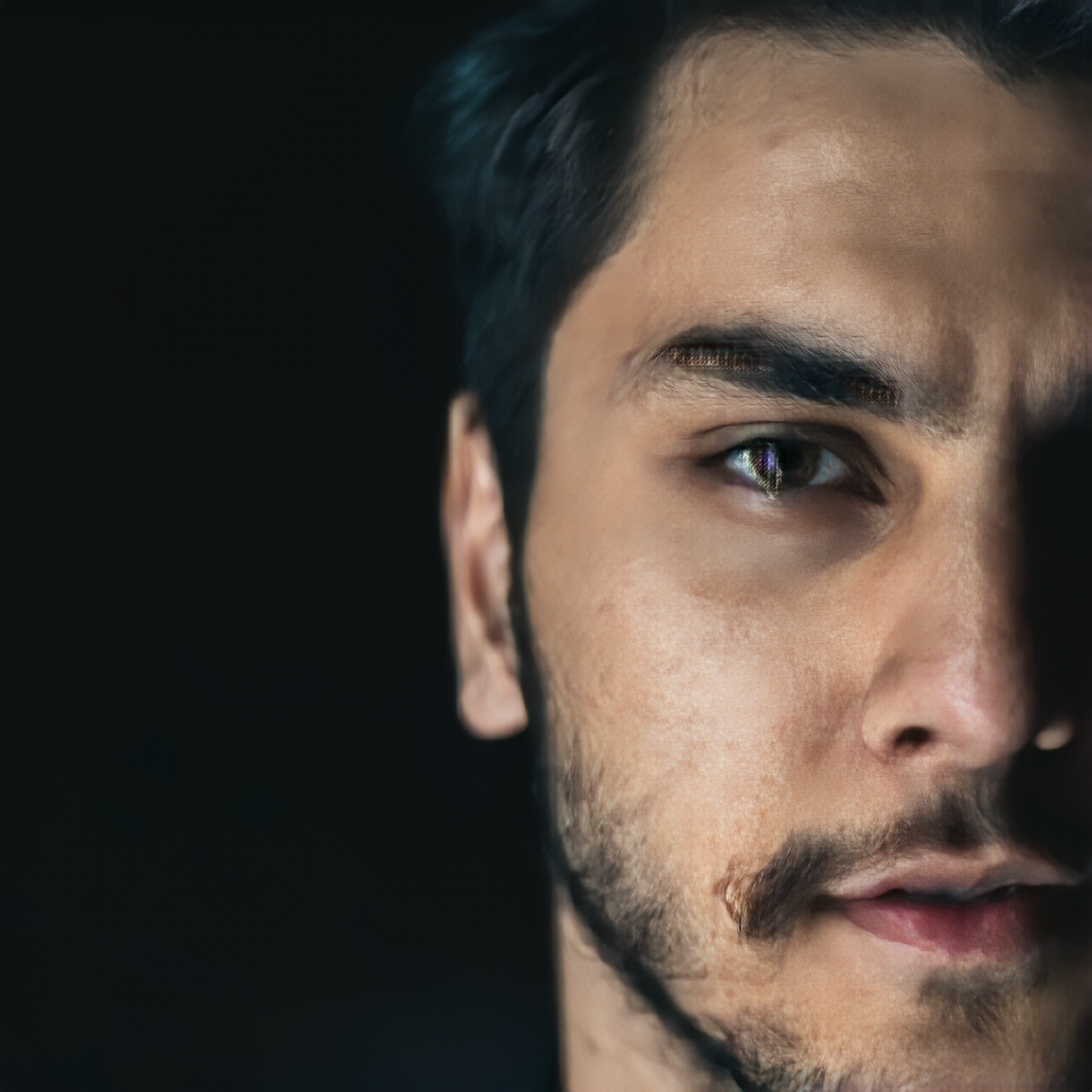}
\end{minipage}
    \caption{Reconstruction Image,iteration 300k, batch size 1, compression rate 1:205, intermediate results}
    \label{fig:enter-label}
\end{figure}
\newpage
\subsection{More samples on CelebA-1024}
\label{sampleceleba}
\begin{figure}[hbp]

\begin{minipage}{\textwidth}
\includegraphics[width =\linewidth]{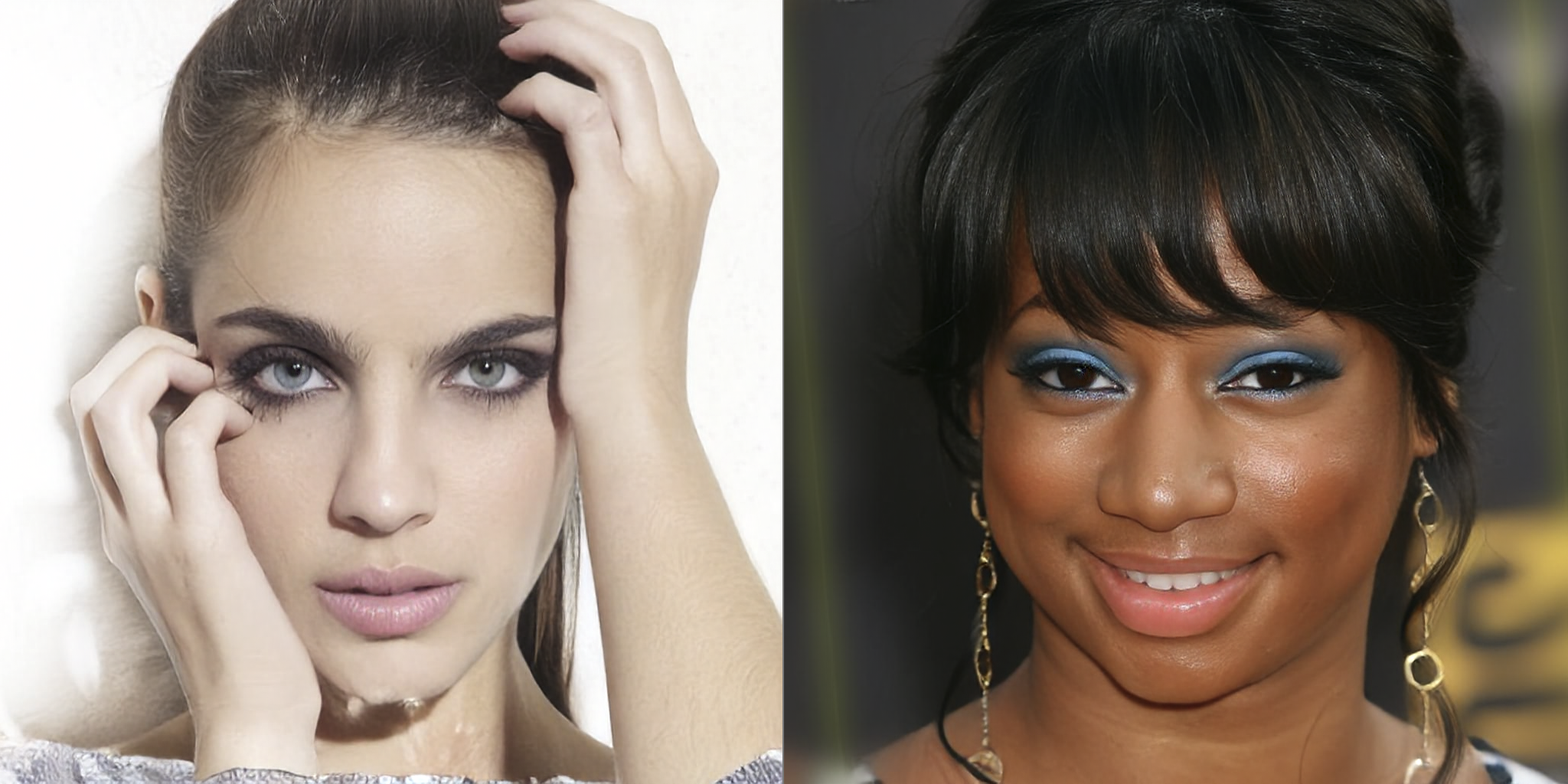}
\end{minipage}
\begin{minipage}{\textwidth}
\includegraphics[width =\linewidth]{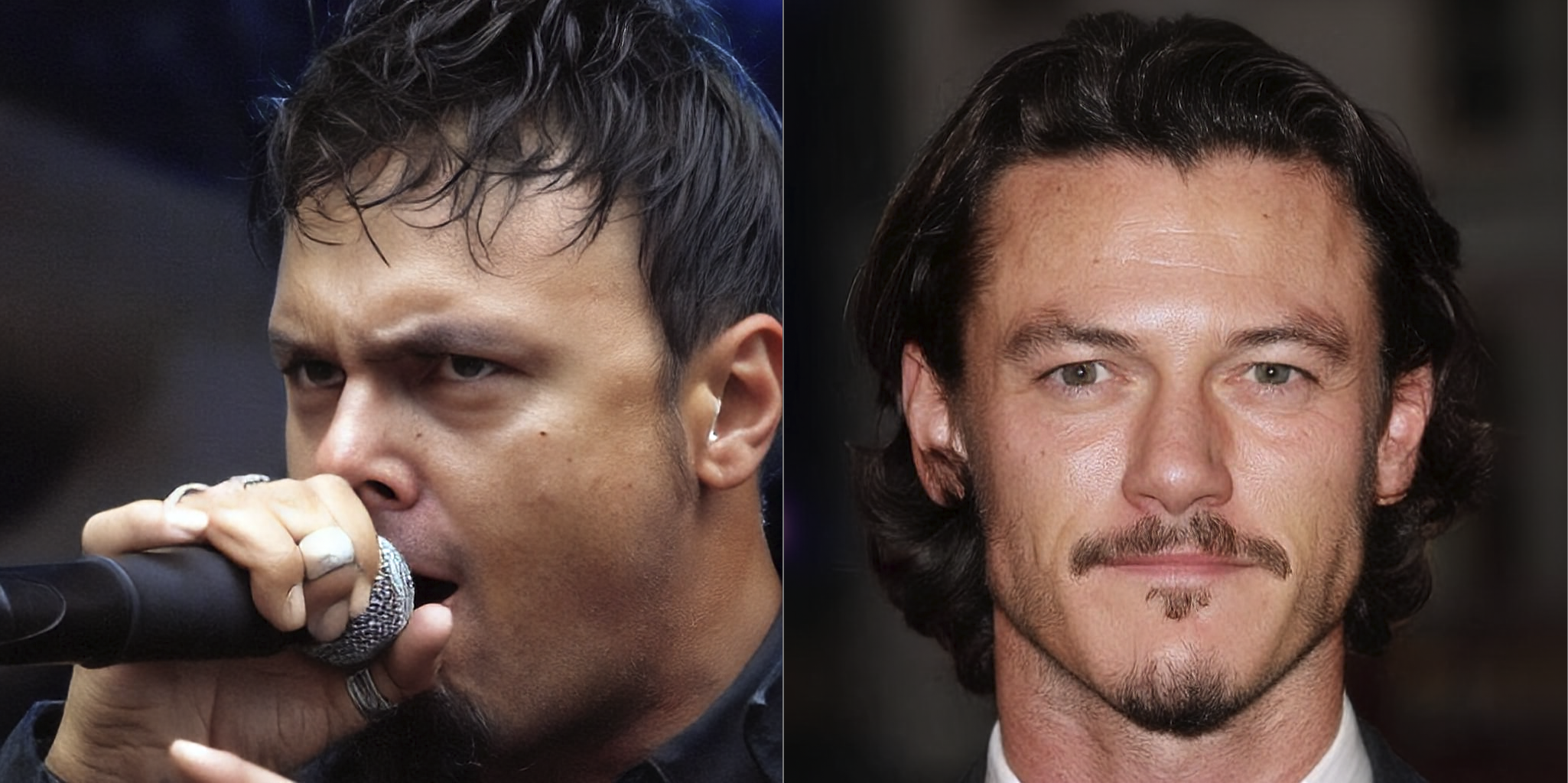}
\end{minipage}
    \caption{Reconstruction Image of CelebaA with 1024*1024 pixels; compression rate 1:256}
    \label{fig:enter-label}
\end{figure}

\begin{figure}[hbp]

\begin{minipage}{.5\textwidth}
\includegraphics[width = \linewidth]{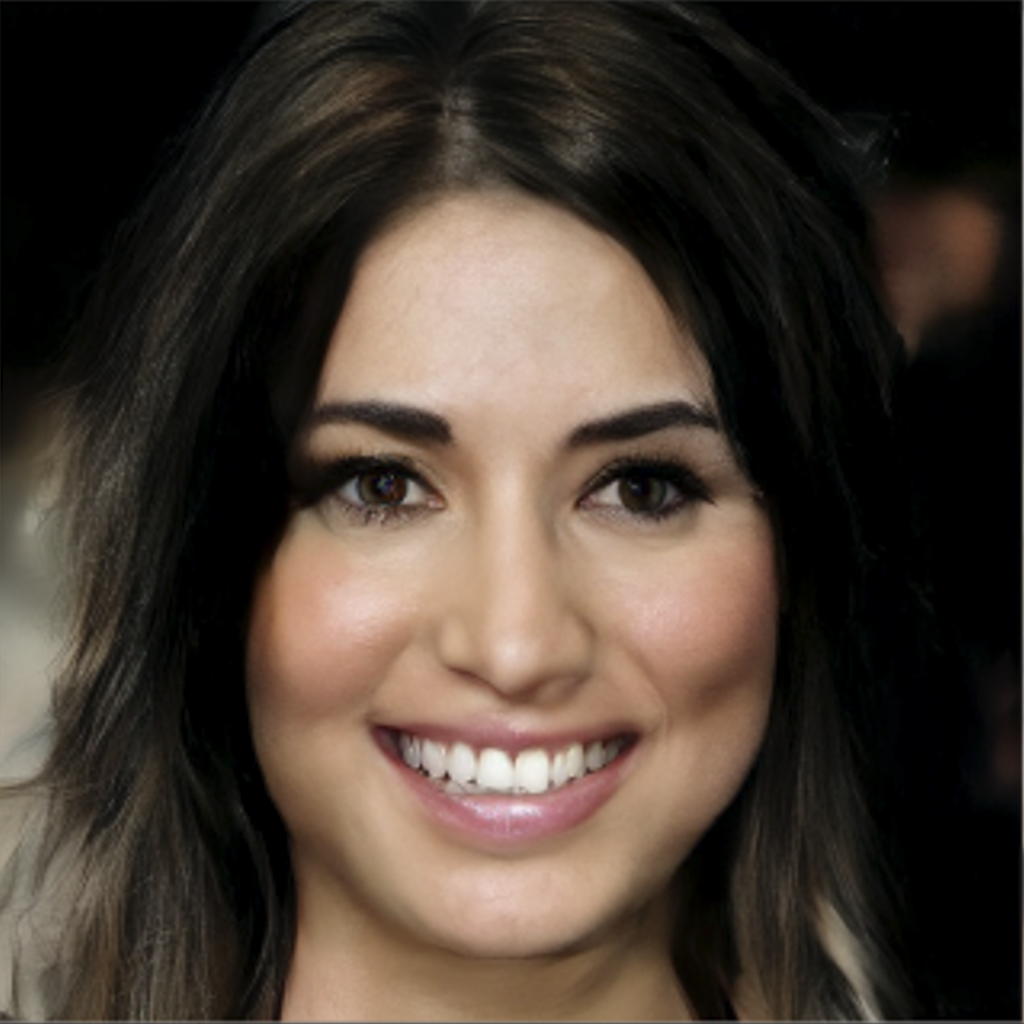}
\end{minipage}
\begin{minipage}{.5\textwidth}
\includegraphics[width =\linewidth]{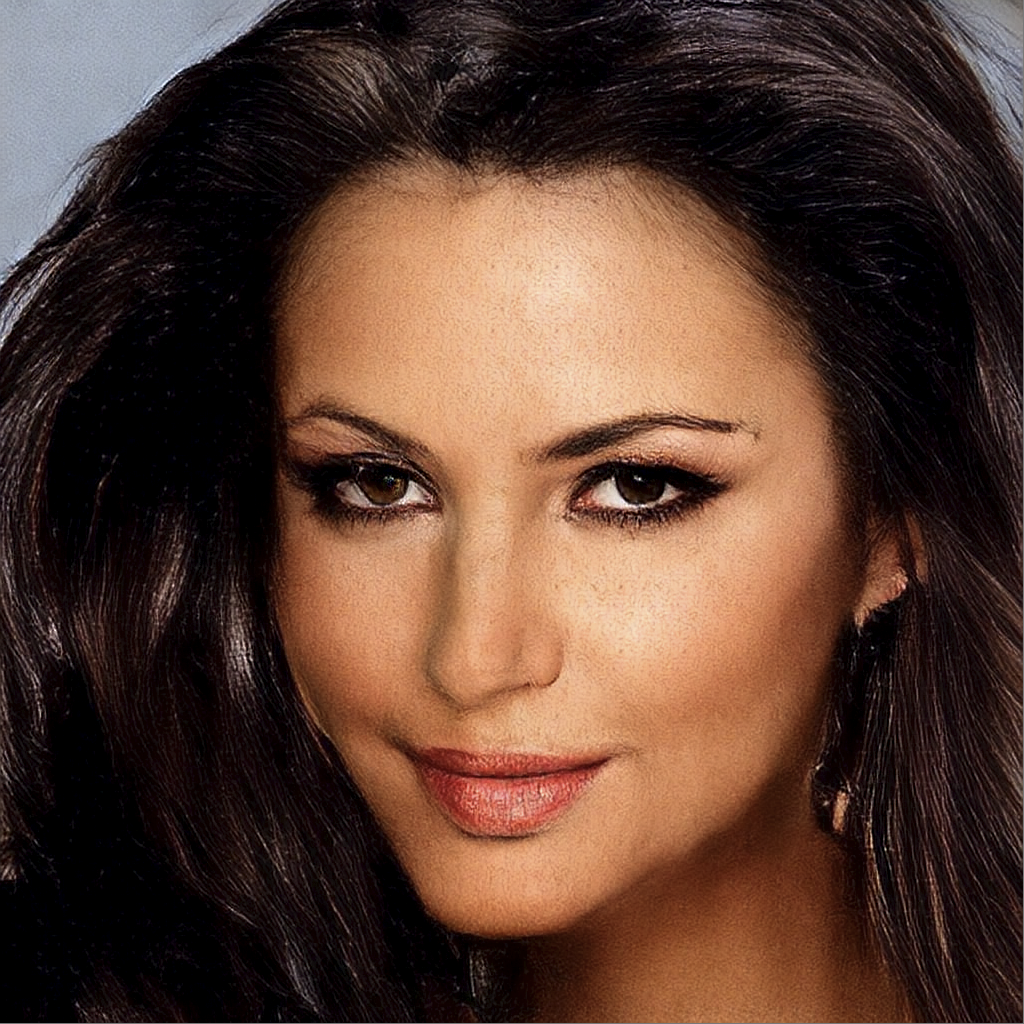}
\end{minipage}
\begin{minipage}{.5\textwidth}
\includegraphics[width = \linewidth]{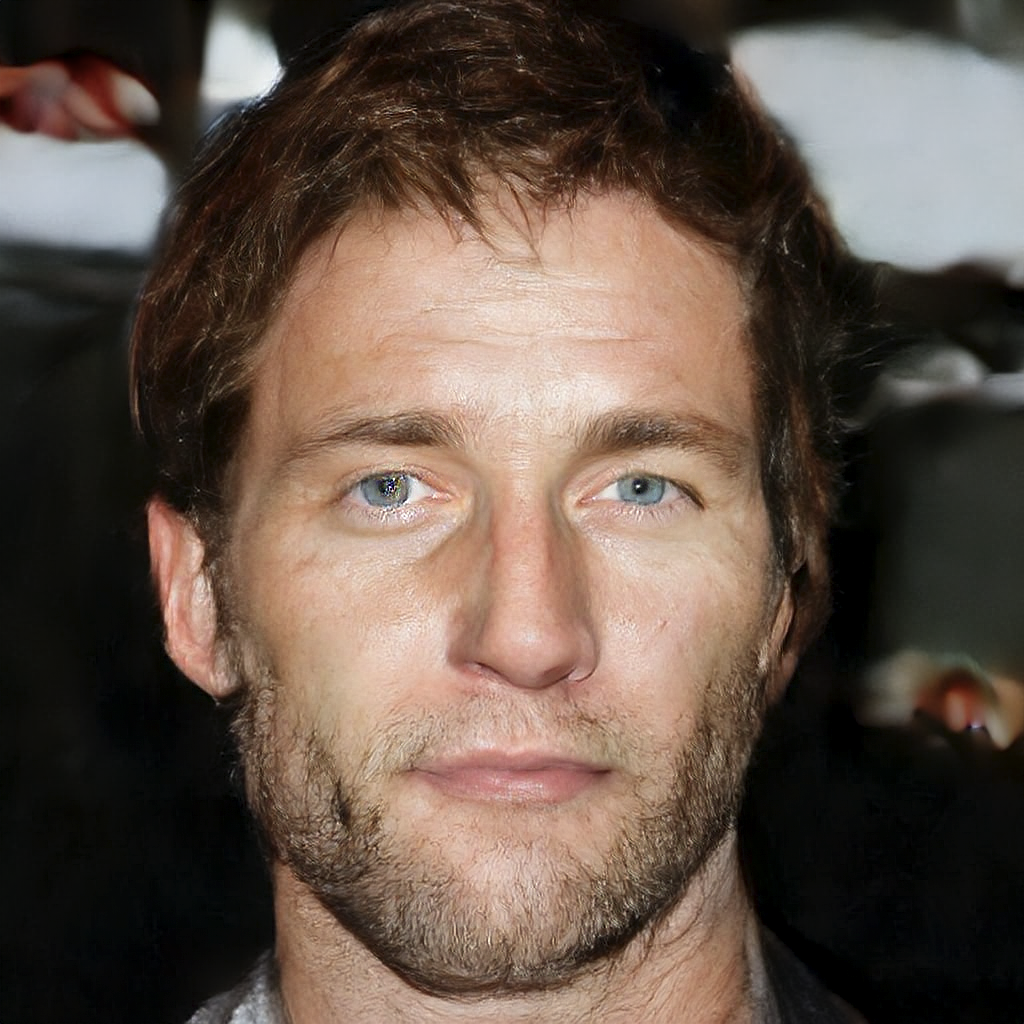}
\end{minipage}
\begin{minipage}{.5\textwidth}
\includegraphics[width =\linewidth]{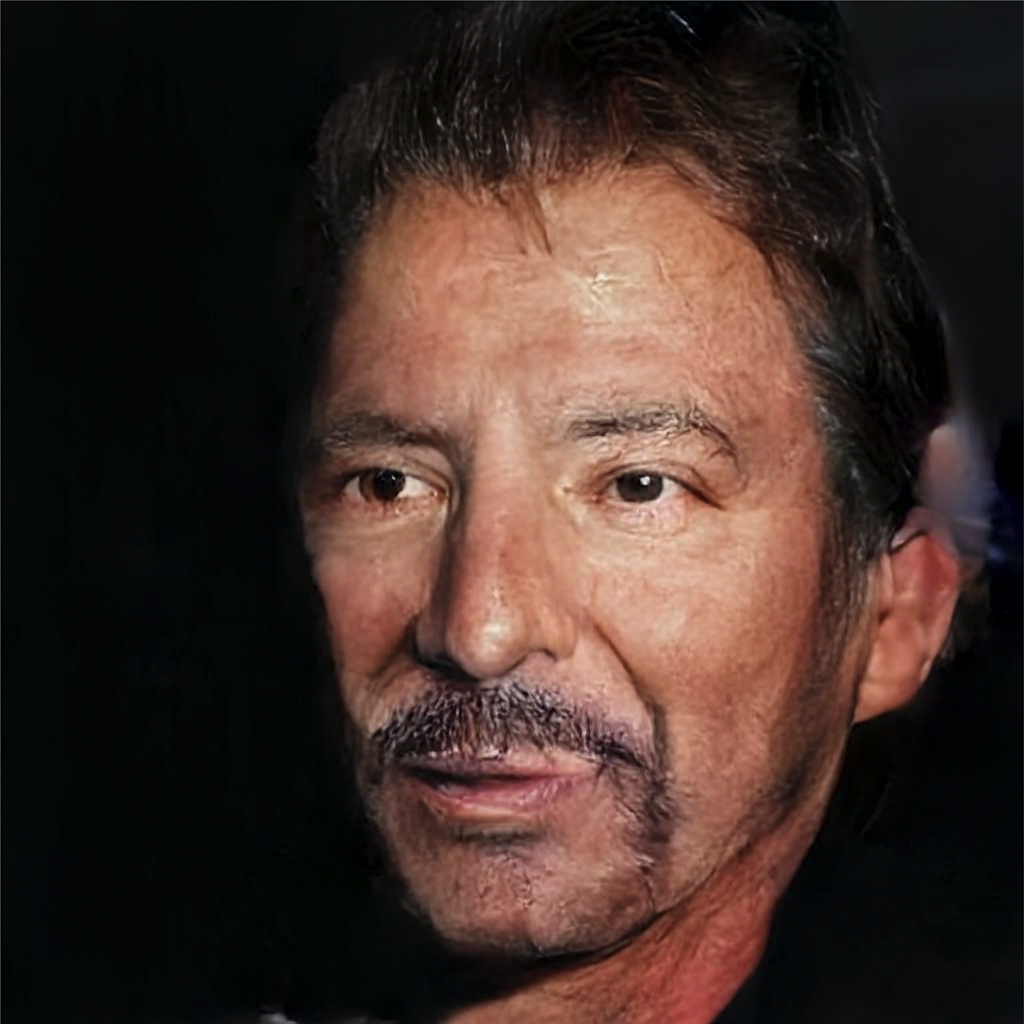}
\end{minipage}
    \caption{Generated Image, sampling with RK45, intermediate results}
    \label{fig:enter-label}
\end{figure}

\subsection{More samples on LSUN-Bedrooms}

\end{document}